\documentclass[journal,10pt,onecolumn]{IEEEtran}
\usepackage{cite}
\usepackage{color,yfonts}
\usepackage[pdftex]{graphicx}
\graphicspath{{fig/}{jpeg/}}
\pdfoutput=1
\usepackage[cmex10]{amsmath}
\usepackage{amssymb}
\usepackage{algorithm}
\usepackage{algpseudocode}
\usepackage{amsmath,amssymb,lipsum}
\usepackage{hyperref}
\usepackage{comment}
\usepackage{pgf} \usepackage{pgfplots} \usepackage{tikz}\usetikzlibrary{shapes,arrows,automata}
\usepackage{graphicx}
\usepackage[font=small]{caption}
\usepackage{subcaption}
\usepackage{mathtools,enumerate}
\input{mysymbol.sty}
\usepackage{needspace}

\usepackage{theorem}
\newtheorem{thm}{Theorem}
\newtheorem{lemma}{Lemma}
\newtheorem{proposition}{Proposition}

\newtheorem{example}{\bf Example}
\theoremstyle{definition}
\newtheorem{assumption}{Assumption}
\newtheorem{remark}{Remark}

\def \QED {$\blacksquare$}

\def \RV  {\text{RV}}
\DeclareMathOperator{\sgn}{sgn}

\title{D4L: Decentralized Dynamic Discriminative Dictionary Learning }
\author{Alec Koppel$^\star$, Garrett Warnell$^{\dagger}$, Ethan Stump$^{\dagger}$, and Alejandro Ribeiro$^\star$
\thanks{Work in this paper is supported by NSF CCF-1017454, NSF CCF-0952867, ONR N00014-12-1-0997, ARL MAST CTA, and ASEE SMART. Part of the results in this paper appeared in \cite{cKoppelEtal15a,cKoppelEtal15b}.}
\thanks{ \noindent$^\star$Department of Electrical and Systems Engineering, University of Pennsylvania, 200 South 33rd Street, Philadelphia, PA 19104. Email: \{akoppel, aribeiro\}@seas.upenn.edu.}
\thanks{\noindent $^{\dagger}$U.S. Army Research Laboratory, Computation and Information Sciences Directorate, 2800 Powder Mill Road, Adelphi, MD 20783. Email: \{garrett.a.warnell.civm, ethan.a.stump2.civ\}@mail.mil} \
}

\begin{document}

\maketitle
\thispagestyle{empty}
\pagestyle{empty}

\begin{abstract}
We consider discriminative dictionary learning in a distributed online setting, where a network of agents aims to learn a common set of dictionary elements of a feature space and model parameters while sequentially receiving observations. We formulate this problem as a distributed stochastic program with a non-convex objective and present a block variant of the Arrow-Hurwicz saddle point algorithm to solve it. Using Lagrange multipliers to penalize the discrepancy between them, only neighboring nodes exchange model information.
 We show that decisions made with this saddle point algorithm asymptotically achieve a first-order stationarity condition on average. The learning rate depends on the signal source, network, and discriminative task. We illustrate the algorithm performance for solving a large-scale image classification task on a network of interconnected servers and observe that practical performance is comparable to a centralized approach. We apply this method to the problem of a robotic team seeking to autonomously navigate in an unknown environment by predicting unexpected maneuvers, demonstrating the proposed algorithm's utility in a field setting.
\end{abstract}


%
\vspace{-2mm}
\section{Introduction}\label{sec:intro}

We develop a framework to solve machine learning problems in cases where latent geometric structure in the feature space may be exploited. We consider cases where the number of training examples is either very large, or signals are sequentially observed by a platform operating in real-time such as an autonomous robot. In the former case, since the sample size is large-scale, processing a few training examples at a time is necessary due to computational cost. However, doing so at a centralized location may be impractical, which motivates the use of learning techniques that may be done collaboratively by a network of interconnected computing servers. 
In the later case, an autonomous robot with no priors on its operating environment only has access to information based on the path it has traversed, which may omit regions of the feature space crucial for tasks such as learning-based control. By communicating with other robots in a network, individuals may learn over a broader domain associated with that which has been explored by the whole network, and thus more effectively solve autonomous learning tasks.
The problem formulation breaks down into three aspects: developing data-driven feature representations, learning task-driven model parameters over these representations, and extending this problem to dynamic, networked settings.

Formally, consider the problem of computing an alternative representation of a set of vectors where this alternative representation may reveal latent relationships between them.  Broadly, this problem is referred to as unsupervised learning, and techniques designed to address it have yielded important advances in a variety of signal processing applications \cite{Chen98atomicdecomposition}. To learn such a representation, a variety of objectives may be considered. If the vector's dimension is very large, dimensionality reduction is of interest, whereby one aims to find a representation that explains the most data variability across a feature space. Classically this task been approached with principle component analysis \cite{Jolliffe:1986}, which requires orthogonality of the basis elements. Alternatively, if specialized domain knowledge is available, finding representations based on particularized functions, i.e. wavelets for natural imagery \cite{Mallat:2008:WTS:1525499}, is more appropriate. A more general approach to seeking signal representations of a feature space is to learn the basis elements from data, as in dictionary learning. Dictionary learning has been successfully applied to signal reconstruction tasks such as inpainting or denoising \cite{Elad:2006:IDV:2319066.2321231, Mairal07sparserepresentation, Zibulevsky:2001:BSS:1120451.1120461}, and higher level signal processing tasks such as classification \cite{Raina:2007:SLT:1273496.1273592, Wright:2009:RFR:1495801.1496037}.

An important question recently posed in \cite{10.1109/TPAMI.2011.156} is {\it why} one aims to learn a signal representation from data, if not for feeding into a higher-level signal processing task. Thus the authors in  \cite{10.1109/TPAMI.2011.156} propose tailoring the dictionary to a discriminative modeling task, referred to as {\it discriminative} dictionary learning. Such methods have recently shown promise as compared to their unsupervised counterparts \cite{conf/cvpr/MairalBPSZ08, Bradley_2008_6192, conf/cvpr/JiangLD11}.
The problem of developing a dictionary representation of a signal specifically suited to the learning problem of interest is a difficult optimization problem. In the centralized offline setting, this class of problems has been solved with block coordinate descent \cite{Tseng01convergenceof, DBLP:journals/npsc/BezdekH03}, or alternating gradient methods \cite{xu2014globally}; however, these techniques are only effective when the training set is static and not too large. In the centralized online setting, prior approaches have made use of stochastic approximation methods \cite{Elad:2006:IDV:2319066.2321231,Mairal:2010:OLM:1756006.1756008}. 

In this paper, we extend the online discriminative dictionary learning formulation of \cite{10.1109/TPAMI.2011.156}  to networked settings, where a team of agents seeks to learn a common dictionary and model parameters based upon local dynamic information. To do so, we consider tools from stochastic approximation \cite{robbins1951} and its decentralized extensions which have incorporated ideas from distributed optimization such as weighted averaging \cite{Nedic09subgradientmethods, DBLP:journals/corr/abs-1112-2972,RamNedicVeeravalli, yuan2013convergence,chainais2013learning},  dual reformulations where each agent ascends in the dual domain \cite{1506308, JakubiecEtal13}, and primal-dual methods which combine primal descent with dual ascent \cite{ArrHurUza58, Nedic09subgradientmethods,DBLP:dblp_conf/icassp/KoppelJR14,Koppelelat14}. 

Our main technical contribution is the formulation of the dynamic multi-agent discriminative dictionary learning problem as a distributed stochastic program, and the development of a block variant of the primal-dual algorithm proposed in \cite{Koppelelat14}. Moreover, we establish that the proposed method converges in expectation to a first-order stationary solution of the problem. We describe the discriminative dictionary learning and sparse representation problem in Section \ref{sec:prob}. We extend this problem to multi-agent settings, and derive an algorithmic solution which is a block variant of the saddle point algorithm of Arrow and Hurwicz \cite{ArrHurUza58, Koppelelat14} in Section \ref{sec:alg}. In Section \ref{sec:results},  we establish the convergence properties of the method.
In Section \ref{sec:experiments}, we analyze the proposed framework's empirical performance on a texture classification problem based upon image data for a variety of network settings and demonstrate its capacity to solve a new class of collaborative multi-class classification problems in decentralized settings. In Section \ref{sec:robots} we consider the algorithm's use in a mobile robotic team for navigability assessment.
We conclude in Section \ref{sec:conclusion}.
%
\vspace{-.773mm}
\section{Discriminative Dictionary Learning}\label{sec:prob}

Consider a set of $T$ signals $\{\bbx_t\}_{t=1}^T$  each of which lives in an $m$-dimensional feature space so that we have $\bbx_t\in \ccalX \subset \reals^m$. We aim to represent the signals $\bbx_t$ as combinations of a common set of $k$ linear basis elements $\{\bbd_l\}_{l=1}^k$, which are unknown and must also be learned from the data. We group these $k$ basis elements into a dictionary matrix $\bbD=[\bbd_1,\ldots,\bbd_k]\in\reals^{m\times k}$ and denote the coding of $\bbx_t$ as $\bbalpha_t \in \reals^k$. For a given dictionary, the coding problem calls for finding a representation $\bbalpha_t$ such that the signal  $\bbx_t$ is close to its dictionary representation $\bbD\bbalpha_t$. This goal can be mathematically formulated by introducing a loss function $f(\bbalpha_t, \bbD ; \bbx_t)$ that depends on the proximity between $\bbD\bbalpha_t$ and the data point $\bbx_t$ and formulating the coding problem as \cite{aharon2006svd}
\begin{equation} \label{eq:projection}
    \bbalpha^*(\bbD;\bbx_t) 
          := \argmin_{\bbalpha_t\in\reals^k}
                 f (\bbalpha_t, \bbD; \bbx_t) \; .
\end{equation}
Hereafter, we assume that basis elements are normalized to have norms $\|\bbd_l\| \leq 1$ so that the dictionary is restricted to the convex compact set $\ccalD:=\{\bbD\in\reals^{m\times k} : \|\bbd_l\| \leq 1,\ \text{for all\ } l \}$. 

The dictionary learning problem associated with the loss function $f(\bbalpha_t, \bbD ; \bbx_t)$ entails finding a dictionary $\bbD$ such that the signals $\bbx_t$ are close to their representations $\bbD\bbalpha^*(\bbD;\bbx_t)$ for all possible $t$. Here, however, we focus on discriminative problems where the goal is to find a dictionary that is well adapted to a specific classification or regression task \cite{10.1109/TPAMI.2011.156}. Formally, we associate with each $\bbx_t$ a variable $\bby_t\in\ccalY$ that represents a discrete label -- in the case of classification problems -- or a set of associated vectors $\ccalY\subset \reals^q$ -- in the case of regression. We then use the coding $\bbalpha^*(\bbD;\bbx)$ in \eqref{eq:elastic_net} as a feature representation of the signal $\bbx_t$ and introduce the classifier $\bbw$ that is used to predict the label or vector $\bby_t$ when given the signal $\bbalpha^*(\bbD;\bbx)$. The merit of the classifier $\bbw$ is measured by the smooth loss function $h\big(\bbalpha^*(\bbD;\bbx_t), \bbD, \bbw ; (\bbx_t, \bby_t)\big)$ that captures how well the classifier $\bbw$ may predict $\bby_t$ when given the sparse coding $\bbalpha^*(\bbD;\bbx_t)$ that we compute using the dictionary $\bbD$. The discriminative dictionary learning problem is formulated as the joint determination of the dictionary $\bbD\in\ccalD$ and classifier $\bbw\in\ccalW$ that minimize the cost $h\big(\bbalpha^*(\bbD;\bbx_t), \bbw ; (\bbx_t, \bby_t)\big)$ averaged over the  training set,
\begin{equation} \label{eq:batch_disc_dict_prelim}
   (\bbD^*, \bbw^*) 
      := \!\argmin_{\bbD\in\ccalD, \bbw\in\ccalW}
          \!  \frac{1}{T} \sum_{t=1}^T \!
            h \big(\bbalpha^*(\bbD;\bbx_t), \bbD, \bbw ; (\bbx_t, \bby_t)\big). 
\end{equation}
For given dictionary $\bbD$ and signal sample $\bbx_t$ we compute the code $\bbalpha^*(\bbD;\bbx_t)$ as per \eqref{eq:projection}, predict $\bby_t$ using $\bbw$, and measure the prediction error with the loss function $h \big(\bbalpha^*(\bbD;\bbx_t), \bbw ; (\bbx_t, \bby_t)\big)$. The optimal pair $(\bbD^*, \bbw^*)$ in \eqref{eq:batch_disc_dict_prelim} is the one that minimizes the cost averaged over the given sample pairs $(\bbx_t, \bby_t)$. Observe that $\bbalpha^*(\bbD;\bbx_t)$ is not a variable in the optimization in \eqref{eq:batch_disc_dict_prelim} but a mapping for a implicit dependence of the loss on the dictionary $\bbD$. To simplify notation we henceforth write \eqref{eq:batch_disc_dict_prelim} as
\begin{equation} \label{eq:batch_disc_dict}
   (\bbD^*, \bbw^*) 
      := \argmin_{\bbD\in\ccalD, \bbw\in\ccalW}
            \frac{1}{T} \sum_{t=1}^T 
            h \big(\bbD, \bbw ; (\bbx_t, \bby_t) \big) \; . 
\end{equation}
The optimization problem in \eqref{eq:batch_disc_dict} is not assumed to be convex  -- this would be too restrictive because the dependence of $h$ on $\bbD$ is, partly, through the mapping $\bbalpha^*(\bbD;\bbx_t)$ defined by \eqref{eq:projection}. In general, only local minima of \eqref{eq:batch_disc_dict} can be found. 

Our goal in this paper is to study online algorithms that solve \eqref{eq:batch_disc_dict} as training pairs $(\bbx_t, \bby_t)$ become available. To do so we introduce the assumption that training pairs $(\bbx_t, \bby_t)$ are independently sampled from a common probability distribution and replace \eqref{eq:batch_disc_dict} by
\begin{equation} \label{eq:central_disc_dict}
   (\bbD^*, \bbw^*) 
      := \argmin_{\bbD\in\ccalD, \bbw\in\ccalW}
         {\mathbb E}_{\bbx,\bby}
         \Big[h \big(\bbD, \bbw; (\bbx, \bby)\Big].
\end{equation}
The problems in \eqref{eq:central_disc_dict} and \eqref{eq:batch_disc_dict} are equivalent in the limit of $T\to\infty$ if $(\bbx_t, \bby_t)$ are independently drawn from the joint distribution of the random pair $(\bbx,\bby)$. The problem in \eqref{eq:central_disc_dict}, as the one in \eqref{eq:batch_disc_dict}, is not convex. We clarify the problem formulation in \eqref{eq:central_disc_dict} with two representative examples.

%
\begin{example}[Sparse non-discriminative learning]\normalfont
When we have $k<m$, the formulation in \eqref{eq:projection} aims at finding a dictionary that reduces data dimensionality from $m$ to $k$. In this paper we are more interested in the overdetermined case in which $k>m$ but we want the codes $\bbalpha_t$ to be sparse. These sparsity constraints can be written as upper limits on the zero norm of $\bbalpha_t$ but that would yield computationally intractable formulations. To circumvent this issue, sparsity can be incentivized by adding, e.g., elastic net regularization terms \cite{Jenatton2009,bach2011convex}, in which case we can write the loss functions $f (\bbalpha_t, \bbD; \bbx_t)$ in \eqref{eq:projection} as 
\begin{equation}\label{eq:elastic_net}
    f (\bbalpha_t, \bbD ; \bbx_t) =
            \tdf (\bbalpha_t, \bbD; \bbx_t)   
                  + \zeta_1 \|\bbalpha_t\|_1+\frac{\zeta_2}{2} \|\bbalpha_t\|_2^2 \; .
\end{equation}
In \eqref{eq:elastic_net}, $\tdf (\bbalpha_t, \bbD; \bbx_t)$ measures proximity between $\bbx_t$ and $\bbD\bbalpha_t$, the $\ell_1$ term $\zeta_1 \|\bbalpha_t\|_1$ encourages sparsity, and the $\ell_2$ term $ (\zeta_2/2)\|\bbalpha_t\|_2^2$ is a smooth regularizer. Common choices for the proximity functions are the Euclidean distance $\tdf (\bbalpha_t, \bbD; \bbx_t) =\|\bbx_t - \bbD \bbalpha_t \|/2$ and the $l_{\infty}$ norm  $\tdf (\bbalpha_t, \bbD; \bbx_t) = \|\bbx_t - \bbD \bbalpha_t \|_{\infty}=\max_i | \bbx_{i,t} - \bbD_i \bbalpha_{i,t} |$. In a non-discriminative problem we simply want to make $\bbx_t$ and $\bbD\bbalpha^*(\bbD;\bbx_t)$ close to each other across elements of the training set. We achieve that by simply making $h \big(\bbalpha^*(\bbD;\bbx_t), \bbD, \bbw ; (\bbx_t, \bby_t)\big) = f \big(\bbalpha^*(\bbD;\bbx_t), \bbD ; (\bbx_t, \bby_t)\big)$.
\end{example}

%
\begin{example}[Sparse logistic regression] \normalfont Given a training set of pairs $(\bbx_{t}, y_{t})$ where $\bbx_{t}\in\reals^p$ is a feature vector with associated binary label $y_{t} \in \{-1,1\}$, we seek a decision hyperplane $\bbw \in \reals^k$ which best separates data points with distinct labels. However, instead of looking for linear separation in the original space, we seek for linear separation in a sparse coded space. Thus, let $\bbalpha^*(\bbD;\bbx_t)$ be the sparse coding of $\bbx_t$ computed through \eqref{eq:projection} when using the loss function in \eqref{eq:elastic_net}. We want to find a classifier $\bbw$ such that $\bbw^T \bbalpha^*(\bbD;\bbx_t)>0$ when $y_t=1$ and $\bbw^T \bbalpha^*(\bbD;\bbx_t)<0$ when $y_t=0$. This hyperplane need not exist but we can always model the probability of observing $y_t = 1$ through its odds ratio relative to $y_t=-1$. This yields the optimal classifier $\bbw^*$ as the one that minimizes the  logistic loss
\begin{align} \label{eq:logistic_loss}
  h (\bbD, \bbw ; (\bbx_t, \bby_t) )
  		   &= \Pc{y_t = \pm 1 \given \bbalpha^*(\bbD;\bbx_t), \bbw } \nonumber\\
  	       &= \frac{1}{1 + \exp(-y_t \bbw^T \bbalpha^*(\bbD;\bbx_t))} \; .
\end{align}
For a feature vector $\bbx_t$, \eqref{eq:logistic_loss} models the probability of the label $y_{t}$ being $1$ or $-1$ as determined by the inner product $\bbw^T\bbalpha^*(\bbD;\bbx_t)$ through the given logistic transformation. Substituting \eqref{eq:logistic_loss} into \eqref{eq:central_disc_dict} yields the discriminative dictionary learning problem for logistic regression with sparse features.
\end{example}

%
\vspace{-2mm}
\subsection{Decentralized discriminative learning}\label{subsec:new_prob}

We want to solve \eqref{eq:central_disc_dict} in distributed settings where signal and observation pairs are independently observed by agents of a network. Agents aim to learn a dictionary and model parameters that are common with all others while having access to local information only. In particular, associated with each agent $i$ is a local random variable and associated output variable $(\bbx_i,\bby_i)$ and each agent's goal is to learn over the aggregate training domain $\{(\bbx_i,\bby_i)\}_{i=1}^N$. Let $\ccalG=(V,\ccalE)$ be a symmetric and connected network with node set $V=\{1,\ldots,N\}$ and $M=|\ccalE|$ directed edges of the form $e=(i,j)$ and further define the neighborhood of $i$ as the set of nodes $n_i := \{j:(i,j)\in\ccalE\}$ that share an edge with $i$. When each of the $N$ agents observes a pair $(\bbx_i,\bby_i)$, the function $h$ in \eqref{eq:central_disc_dict} can be written as a sum of local losses,
\begin{equation}\label{eq:costsum}
   h (\bbD, \bbw ; (\bbx, \bby) )
      = \sum_{i=1}^N h_i (\bbD_i, \bbw_i ; (\bbx_i, \bby_i) ) ,
\end{equation}
where we have defined the vertically concatenated dictionary $\bbD:=[\bbD_{1};\ldots;\bbD_{N}]\in\reals^{Nm \times k}$ and model parameter $\bbw : =[\bbw_{1};\ldots;\bbw_{N}]\in\reals^{N k}$. 

Substituting \eqref{eq:costsum} into the objective in \eqref{eq:central_disc_dict} yields a problem in which the agents learn dictionaries and classifiers that depend on their local observations only. The problem to be formulated here is one in which the agents learn common dictionaries $\bbD_{i}=\bbD_j$ and models $\bbw_i=\bbw_j$. Since the network $\ccalG$ is assumed to be connected, this relationship can be attained by imposing the constraints $\bbD_{i}=\bbD_{j}$ and $\bbw_{i}=\bbw_{j}$ for all pairs of neighboring nodes $(i,j)\in\ccalE$. Substituting \eqref{eq:costsum} into the objective in  \eqref{eq:central_disc_dict} with these constraints yields the distributed stochastic program
\begin{align} \label{eq:dist_stoch_dict}
   \{\bbD_i^*, \bbw_i^*\}_{i=1}^N :=
   &  \argmin_{\bbD_i\in\ccalD, \bbw_i\in\ccalW} \!\!\!	
   && \sum_{i=1}^N{\mathbb E}_{\bby_i,\bbx_i}
      \Big[h_i (\bbD_i, \bbw_i ; (\bbx_i, \bby_i)\Big] \nonumber\\
   &  \st
   && \bbD_i=\bbD_j,\ \bbw_{i}=\bbw_{j},\ j\in n_i  .
\end{align}
When the agreement constraints in \eqref{eq:dist_stoch_dict} are satisfied, the objective is equivalent to one in which all the observations are made at a central location and a single dictionary and model are learnt. Thus, \eqref{eq:dist_stoch_dict} corresponds to a problem in which each agent $i$, having only observed the local pairs $(\bbx_i,\bby_i)$, aims to learn a dictionary representation and model parameters that are optimal when information is aggregated globally over the network. The decentralized discriminative learning problem is to develop an iterative algorithm that relies on communication with neighbors only so that agent $i$ learns the optimal (common) dictionary $\bbD_i^*=\bbD_j^*$ and discriminative model $\bbw_i^*=\bbw_j^*$. We present in the following section an algorithm that is shown in Section \ref{sec:results} to converge to a local optimum of \eqref{eq:dist_stoch_dict}.  
\vspace{-2mm}
\begin{remark}\label{remark1} \normalfont
Decentralized learning techniques may be applied to solving pattern recognition tasks in networks of autonomous robots operating in real-time, provided that realizations of the output variables are generated by a process which is internal to the individual platforms. In particular, consider the formulation in \eqref{eq:dist_stoch_dict}, and let $\bby_i$ represent the difference between state information associated with a commanded trajectory and that which is observed by on-board sensors of robot $i$. Most robots are equipped with sensors  such as gyroscopes, accelerometers, and inertial measurement units, which make state information available. 

In this case, the interconnected network of robots does not need external supervision or human in the loop in order to perform discriminative learning. In Section \ref{sec:robots} we propose solving problems of the form  \eqref{eq:dist_stoch_dict} in a network of interconnected robots operating in a field setting by generating binary labels $\bby_i\in \{-1,1\}$ by thresholding the difference between measurements made via on-board inertial measurement units (IMU) and movements which are executed in an open-loop manner with a joystick. 

%
\end{remark}

\vspace{-5mm}
\section{Block Saddle Point Method}\label{sec:alg}

To write the constraints in \eqref{eq:dist_stoch_dict} more compactly, define the augmented graph edge incidence matrix $\bbC_D \in \reals^{  (M  \times N) m}$ associated with the dictionary constraint. The matrix $\bbC_D$ is formed by $M\times N$ square blocks of dimension $m$. If the edge $e=(i,j)$ links node $i$ to node $j$ the block $(e,i)$ is $[\bbC_D]_{ei}=\bbI_{m}$ and the block $[\bbC_D]_{ej}=-\bbI_{m}$, where $\bbI_{m}$ denotes the identity matrix of dimension $m$. All other blocks are identically null, i.e., $[\bbC]_{ei}=[\bbC]_{ej}=\bbzero_{m\times m}$ when $e\neq (i,j)$. Likewise, the matrix $\bbC_w\in\reals^{(M \times N )k}$ is defined by $M \times N $ blocks of dimension $k$ with $[\bbC_w]_{ei}=\bbI_{k}$ and $[\bbC_w]_{ej}=-\bbI_{k}$ when $e=(i,j)$ and $[\bbC]_{ei}=[\bbC]_{ej}=\bbzero_{k}$ otherwise. Then the constraints $\bbD_{i}=\bbD_{j}$ and $\bbw_i=\bbw_j$ for all pairs of neighboring nodes can be written as
\begin{align} \label{eq:noco_newconstraint}
    \bbC_D \bbD = \bbzero, \qquad 
    \bbC_w \bbw = \bbzero.
\end{align}
The edge incidence matrices $\bbC_D$ and $\bbC_w$ have exactly $m$ and $k$ null singular values, respectively. We denote as $0 < \gamma$ the smallest nonzero singular value of $\bbC : = [\bbC_D;  \bbC_w]$ and as $\Gamma$ the largest singular value of $\bbC$, which both measure network connectedness. 

Imposing the constraints in \eqref{eq:noco_newconstraint} for all realizations of the local random variables requires global coordination -- indeed, the formulation would be equivalent to the centralized problem in \eqref{eq:central_disc_dict}. Instead, we consider a modification of \eqref{eq:costsum} in which we add linear penalty terms to incentivize the selection of coordinated actions. Introduce then dual variables $\bbLambda_{e} = \bbLambda_{ij} \in\reals^{m\times k}$ associated with the constraint $\bbD_{i}-\bbD_{j}=\bbzero$ and consider the addition of penalty terms of the form $\text{tr}[\bbLambda_{ij}^T(\bbD_{i}-\bbD_{j})]$. For an edge that starts at node $i$, the multiplier $\bbLambda_{ij}$ is assumed to be kept at node $i$. Similarly, introduce dual variables $\bbnu_{ij}$  associated with the constraint  $\bbw_{i}-\bbw_{j}=\bbzero$ for all neighboring node pairs and penalty terms $\bbnu_{ij}^T(\bbw_{i}-\bbw_{j})$. By introducing the stacked matrices $\bbLambda:=[\bbLambda_{1};\ldots;\bbLambda_{M}]\in\reals^{Mm\times k}$ and $\bbnu :=[\bbnu_{1};\ldots;\bbnu_{M}]\in\reals^{M k}$ which are restricted to compact convex sets $\mathfrak{L}$ and $\ccalN$, we can write the Lagrangian of this problem as
\vspace{-1mm}
\begin{align} \label{eq:lagrangian}
   \ccalL(\bbD,\bbw, \bbLam,\bbnu) \ = \ &
	\sum_{i=1}^N{\mathbb E}_{\bby_i,\bbx_i}\left[h_i (\bbD_i, \bbw_i ; (\bbx_i, \bby_i))\right]
	 \nonumber \\
	& \quad+ \text{tr}\big(\bbLambda^T \bbC_D \bbD\big) 
	+ \bbnu^T \bbC_w \bbw \; .
\end{align}

Recall that in the applications considered here the optimization problem in \eqref{eq:dist_stoch_dict} is nonconvex. Thus, we use the dual formulation in \eqref{eq:lagrangian} to develop an iterative distributed algorithm that converges to a KKT point of \eqref{eq:dist_stoch_dict}.

To do so, suppose agent $i$ receives local observation pairs $(\bbx_{i,t}, \bby_{i,t})$ at time $t$ and define the instantaneous Lagrangian as the stochastic approximation of \eqref{eq:lagrangian} evaluated with the observations $\{(\bbx_{i,t}, \bby_{i,t})\}_{i=1}^N$ aggregated across the network as
\begin{align} \label{eq:stoch_lagrangian}
   \hat{\ccalL}_t(\bbD,\bbw, \bbLam,\bbnu) \ = \ &
	\sum_{i=1}^N h_i (\bbD_i, \bbw_i ; (\bbx_{i,t}, \bby_{i,t}))
	 \nonumber \\
	& \quad+ \text{tr}\big(\bbLambda^T \bbC_D \bbD\big) 
	+ \bbnu^T \bbC_w \bbw \; .
\end{align}
We consider the use of the Arrow-Hurwicz saddle point method to solve \eqref{eq:dist_stoch_dict} by alternating block variable updates, in order to exploit the fact that primal-dual stationary pairs are saddle points of the Lagrangian to work through successive primal alternating gradient descent steps and dual gradient ascent steps. Particularized to the stochastic approximate Lagrangian in \eqref{eq:stoch_lagrangian}, the primal iteration of the saddle point algorithm takes the form
\begin{align} \label{eq:sp_dict}
   \bbD_{t+1}   &=  \ccalP_{\ccalD^N}\big[\bbD_t - \eps_t {\nabla}_\bbD \hat{\ccalL}_t (\bbD_t,\bbw_t, \bbLam_t,\bbnu_t) \big]\;, \\
   \label{eq:sp_w}
      \bbw_{t+1}   &=\ccalP _{\ccalW^N}\big[ \bbw_t - \eps_t {\nabla}_\bbw \hat{\ccalL}_t (\bbD_{t},\bbw_t, \bbLam_t,\bbnu_t) \big]_{}\; , 
\end{align}
where  $\tilde{\nabla}_\bbD \hat{\ccalL}_t (\bbD_t,\bbw_t, \bbLam_t,\bbnu_t)$ and $\tilde{\nabla}_\bbw \hat{\ccalL}_t (\bbD_t,\bbw_t, \bbLam_t,\bbnu_t)$, are projected stochastic subgradients of the Lagrangian with respect to $\bbD$ and $\bbw$, respectively. Moreover, $\ccalP_{\ccalD^N}[\cdot]$ and $\ccalP_{\ccalW^N}[\cdot]$ denote orthogonal projections onto feasible sets $\ccalD^N$ and $\ccalW^N$.
%
%
%
Likewise, the dual iteration is defined as
\begin{align} 
   \bbLam_{t+1} &= \ccalP _{\mathfrak{L}^M}\big[\bbLam_t + \eps_t {\nabla}_{\bbLam} \hat{\ccalL}_t (\bbD_{t+1},\bbw_{t+1}, \bbLam_{t},\bbnu_{t})\big]_{} \; , \label{eq:sp_dual_d}
   \\
   \bbnu_{t+1} &= \ccalP _{\ccalN^M} \big[\bbnu_t + \eps_t {\nabla}_{\bbnu} \hat{\ccalL}_t (\bbD_{t+1},\bbw_{t+1}, \bbLam_{t},\bbnu_{t})\big] \; ,\label{eq:sp_dual_w}
\end{align}
where ${\nabla}_{\bbLam} \hat{\ccalL}_t (\bbD_t,\bbw_t, \bbLam_t,\bbnu_t) $ and $ {\nabla}_{\bbnu} \hat{\ccalL} (\bbD_t,\bbw_t, \bbLam_t,\bbnu_t) $ are the stochastic subgradients of the Lagrangian with respect to $\bbLam$ and $\bbnu$, respectively. Moreover, $\ccalP_{\mathfrak{L}^M}[\cdot]$ and $\ccalP_{\ccalN^M}[\cdot]$ denote orthogonal projections onto dual feasible sets $\mathfrak{L}^M$ and $\ccalN^M$ which are compact subsets of Euclidean space in the appropriate dimension.
Additionally, $\eps_t$ is a step size chosen as $O(1/t)$ -- see Section \ref{sec:results}. 

We now show that the algorithm specified by \eqref{eq:sp_dict}-\eqref{eq:sp_dual_w} yields an effective tool for discriminative learning in multi-agent settings.

%
\vspace{-2mm}
\begin{proposition}\label{prop} 
The gradient computations in \eqref{eq:sp_dict}-\eqref{eq:sp_w} may be separated along the local primal variables $\bbD_{i,t}$ and $\bbw_{i,t}$ associated with node $i$, yielding $2N$ parallel updates
 \begin{align}\label{eq:dict_update}
    \bbD_{i,t+1} 
      & =\ccalP_{\ccalD}\Big[   \bbD_{i,t}  - \eps_t  
          \Big(  	 \nabla_{\bbD_i} h_i (\bbD_{i,t}, \bbw_{i,t} ; (\bbx_{i,t}, \bby_{i,t}) ) \nonumber \\
	 & \quad+\tbD_{i,t}+ \sum_{j \in n_i} (\bbLam_{ij,t} -\bbLam_{ji,t}) \Big) \Big]\;, \\
\label{eq:param_update}
   \bbw_{i,t+1} 
      & =  \ccalP_{\ccalW}\Big[\bbw_{i,t} - \eps_t  
         \Big(\nabla_{\bbw_i} h_i (\bbD_{i,t}, \bbw_{i,t} ; (\bbx_{i,t}, \bby_{i,t}) ) \nonumber\\
	 &\qquad+ \sum_{j \in n_i} (\bbnu_{ij,t} -\bbnu_{ji,t})\Big)\Big]\;,
\end{align}
where $\ccalP_{\ccalD}[\cdot]$ denotes the orthogonal projection operator onto set $\ccalD$, and likewise for $\ccalP_{\ccalW}[\cdot]$.
Moreover, the dual gradients in the updates of $\bbLam_{ij,t}$ and $\bbnu_{ij,t}$ respectively in \eqref{eq:sp_dual_d}-\eqref{eq:sp_dual_w}  may separated into $2M$ parallel updates associated with edge $(i,j)$
\begin{align} \label{eq:sda_dual_lam_ij}
   \bbLam_{ij,t+1} &=\ccalP_{\mathfrak{L}}\big[  \bbLam_{ij,t} + \eps_t \left( \bbD_{i,t+1} - \bbD_{j,t+1}  \right)\Big]   \; , \\
\label{eq:sda_dual_nu_ij}
   \bbnu_{ij,t+1}& =\ccalP_{\ccalN}\big[ \bbnu_{ij,t} + \eps_t \left( \bbw_{i,t+1} - \bbw_{j,t+1}  \right) \Big]  \;,
\end{align}
which allows for distributed computation across the network. Again, $\ccalP_{\mathfrak{L}}[\cdot]$ and $\ccalP_{\ccalN}[\cdot]$ denote projections onto sets $\mathfrak{L}$ and $\ccalN$.
\end{proposition}
\vspace{-2mm}
\begin{myproof}
See Appendix A.
\end{myproof}

The D4L algorithm follows by letting node $i$ implement \eqref{eq:dict_update}-\eqref{eq:param_update} as we summarize in Algorithm \ref{alg:d4l}. To do so, node $i$ utilizes its local primal iterates $\bbD_{i,t}$ and $\bbw_{i,t}$, its local dual iterates $\bbLam_{ij}$ and $\bbnu_{ij}$, and its local instantaneous observed pair $(\bbx_{i,t}, \bby_{i,t})$. Node $i$ also needs access to the neighboring multipliers $\bbLam_{ji}$ and $\bbnu_{ji}$ to implement \eqref{eq:dict_update} and \eqref{eq:param_update} as well as to the neighboring primal iterates $\bbD_{j,t}$ and $\bbw_{j,t}$ to implement \eqref{eq:sda_dual_lam_ij} and \eqref{eq:sda_dual_nu_ij}. The core steps of D4L in Algorithm \ref{alg:d4l} are the primal iteration in Step 5 and the dual iteration in Step 7. Steps 4 and 6 refer to the exchange of dual and primal variables that are necessary to implement steps 5 and 7, respectively. Step 1 refers to the acquisition of the signal and observation pair and Step 2 to the computation of the code in \eqref{eq:projection} using the local current dictionary iterate $\bbD_{i,t}$. We discuss the specific use of Algorithm \ref{alg:d4l} to learning discriminative sparse signal representations in a distributed setting to clarify ideas.

%
{\small \begin{algorithm}[t] 
\caption{D4L: Decentralized Dynamic Discriminative Dictionary Learning}
{\small \label{alg:d4l} 
\begin{algorithmic}[1]
\Require $\bbD_0$, initial dictionary, $\bby_u$, local random variables, $\zeta\in\reals$, regularization parameter.
\For {$t=0,1,2,\ldots$}
   \State Acquire local signal and observation pair $(\bbx_{i,t}, \bby_{i,t})$.
   
   \State Coding [cf. \eqref{eq:projection}],   
          $
             \bbalpha^*_{i,t+1}
                := \argmin_{\bbalpha\in\reals^k}
                    f (\bbalpha, \bbD_{i,t}; \bbx_{i,t})  .
          $
   \State Send $(\bbLam_{ij,t}, \bbnu_{ij,t})$ and 
          receive $(\bbLam_{ji,t}, \bbnu_{ji,t})$ for all $j\in n_i$. 
   \State Update dictionary and model parameters 
          [cf. \eqref{eq:dict_update} and \eqref{eq:param_update}]
          \begin{align*}
    \bbD_{i,t+1} 
      & =  \ccalP_{\ccalD}\Big[ \bbD_{i,t} \! -\! \eps_t  
         \Big(  	 \nabla_{\bbD_i} h_i (\bbD_{i,t}, \bbw_{i,t} ; (\bbx_{i,t}, \bby_{i,t}) ) \nonumber \\
	 & \quad+ \sum_{j \in n_i} (\bbLam_{ij,t} -\bbLam_{ji,t}) \Big)\Big] \;, \\
   \bbw_{i,t+1} 
      & = \ccalP_{\ccalW}\Big[ \bbw_{i,t} - \eps_t  
          \Big(\!\nabla_{\bbw_i} h_i (\bbD_i, \bbw_{i,t} ; (\bbx_{i,t}, \bby_{i,t}) ) \nonumber\\
	 &\qquad+ \sum_{j \in n_i} (\bbnu_{ij,t} -\bbnu_{ji,t})\Big)\Big] \;.
\end{align*}
   \State Send $(\bbD_{i,t},\bbw_{i,t})$ and
          receive $(\bbD_{j,t},\bbnu_{j,t})$ for all $j\in n_i$. 
   \State Update Lagrange multipliers 
          [cf. \eqref{eq:sda_dual_lam_ij} and \eqref{eq:sda_dual_nu_ij}]
\begin{align*} 
   \bbLam_{ij,t+1} = \ccalP_{\mathfrak{L}}\Big[ \bbLam_{ij,t} + \eps_t \left( \bbD_{i,t+1} - \bbD_{j,t+1}\right)\Big]  \;,\\
   \bbnu_{ij,t+1} =\ccalP_{\ccalN}\Big[ \bbnu_{ij,t} + \eps_t \left( \bbw_{i,t+1} - \bbw_{j,t+1} \right) \Big]\;.
\end{align*}
\EndFor
\end{algorithmic}}
\end{algorithm}}
%
\begin{example}[Distributed sparse dictionary learning]\normalfont
Consider a multi-agent system in which signals are independently observed at each agent, and the data domain has latent structure which may be revealed via learning discriminative representations that are sparse. In this case, we select the particular form of $f$ in \eqref{eq:projection} as the elastic net [cf. \eqref{eq:elastic_net}] with the Euclidean distance $\tdf (\bbalpha_t, \bbD; \bbx_t) =\|\bbx_t - \bbD \bbalpha_t \|/2$. Then the dictionary update in \eqref{eq:dict_update} may be derived from the subgradient optimality conditions of the elastic-net (see \cite{bach2011convex}): 
\begin{align}\label{eq:elastic_net_opt}
\bbd_l (\bbx_{i,t} - \bbD \bbalpha^*) - \zeta_2 \bbalpha^*_l & = \zeta_1 \sgn(\bbalpha^*_l) &&\text{ if } \bbalpha^*_l \neq 0 \; ,
\nonumber\\
 \bbd_l (\bbx_{i,t} - \bbD \bbalpha^*) - \zeta_2 \bbalpha^*_l & \leq \zeta_1 &&\text{ otherwise} \; ,
\end{align} 
where $\sgn(\bbalpha^*)$ is a vector of signs of $\bbalpha^*$. Proceeding as in the Appendix of \cite{10.1109/TPAMI.2011.156}, define $Z \subset \{1,\dots,k\}$ as the set of nonzero entries of $\bbalpha^*= \bbalpha^*(\bbx,\bbD)$. Then $\bbalpha^*$ is the solution to the system of linear inequalities in \eqref{eq:elastic_net_opt}, i.e.
\begin{align}\label{eq:alpha_tilde}
\bbalpha^*_{Z}= (\bbD_Z^T \bbD_Z + \zeta_2 I)^{-1} (\bbD_Z \bbx - \zeta_1 \sgn(\bbalpha^*)) \; .
\end{align} 
At time $t$,  to compute the stochastic gradient with of \eqref{eq:lagrangian} respect to a local dictionary for agent $i$, apply Proposition 1 of \cite{10.1109/TPAMI.2011.156} which yields the explicit form
\begin{align} \label{eq:dict_grad}
   \nabla_{\bbD_i} \hat{\ccalL}_t (\bbD_t,\bbw_t, \bbLam_t) \!
      &=\!  -\bbD_{i,t} \bbbeta_{i,t} \bbalpha^*_{i,t} 
      + (\bbx_{i,t} \! - \! \bbD_{i,t}\bbalpha^*_{i,t}){\bbbeta_{i,t}}^T \nonumber \\ 
         &\quad   + \tbD_{i,t}+  \sum_{j \in n_i} (\bbLam_{ij,t} -\bbLam_{ji,t}) \; .
\end{align}
$\bbalpha^{*}_{i,t}=\bbalpha^{*}_{i,t}(\bbD_{i,t};\bbx_{i,t})$ is shorthand for \eqref{eq:elastic_net} and $Z_{i,t}$ is defined as the set of indices associated with nonzero entries of $\bbalpha^{*}_{i,t}$. Moreover, we define $\bbbeta^{i,t} \in \reals^k$ as
\begin{align}\label{eq:beta}
\bbbeta^{i,t}_{l}
		&= ([\bbD_{i,t}]_l ^T [\bbD_{i,t}]_l  +  \zeta_2 I)^{-1} \times \nonumber \\
		 &\qquad\qquad \nabla_{\bbalpha_{l}}   h_i (\bbD_{i,t}, \! \bbw_{i,t} ; ( \!\bbx_{i,t}, \bby_{i,t} ) ) \text{ if } l \in Z_{i,t} \; , \nonumber\\
\bbbeta^{i,t}_{l} &=0 \text{ if } l \notin Z_{i,t} \; ,
\end{align} 
as in \cite{10.1109/TPAMI.2011.156}, Proposition 1. This result is established via a perturbation analysis of the elastic-net optimality conditions.  Note that this follows from substituting the solution of \eqref{eq:elastic_net} into $h_i$ and applying the chain rule in the gradient computation. 
\end{example}

\vspace{-3mm}
\section{Convergence Analysis}\label{sec:results}

We turn to establishing that the saddle point algorithm in \eqref{eq:sp_dict}-\eqref{eq:sp_dual_w} asymptotically converges to a stationary point of the problem \eqref{eq:dist_stoch_dict}. Before proceeding with our analysis, we define the primal descent direction with respect to $\bbD$ associated with the projected block stochastic saddle point method as 
\begin{align}\label{eq:proj_grad_def}
\tilde{\nabla}_\bbD \hat{\ccalL}_t& (\bbD_t,\bbw_t, \bbLam_t,\bbnu_t) \\
&=\Big(\bbD_t - \ccalP_{\ccalD^N}\Big[\bbD_t - \eps_t\nabla_\bbD \hat{\ccalL}_t 
(\bbD_t,\bbw_t, \bbLam_t,\bbnu_t) \Big]\Big)/\eps_t \; , \nonumber
\end{align}
and the dual ascent direction with respect to $\bbLambda$ as
\begin{align}\label{eq:proj_grad_def2}
\tilde{\nabla}_{\bbLambda} \hat{\ccalL}_t& (\bbD_{t+1},\bbw_{t+1}, \bbLam_t,\bbnu_t) \\
&=\Big(\bbLambda_t - \ccalP_{\mathfrak{L}^M}\Big[\bbLambda_t + \eps_t\nabla_{\bbLambda} \hat{\ccalL}_t 
(\bbD_{t+1},\bbw_{t+1}, \bbLam_t,\bbnu_t) \Big]\Big)/\eps_t \; . \nonumber
\end{align}
The projected stochastic gradients $\tilde{\nabla}_\bbw \hat{\ccalL}_t (\bbD_t,\bbw_t, \bbLam_t,\bbnu_t) $ and $\tilde{\nabla}_{\bbnu} \hat{\ccalL}_t (\bbD_{t+1},\bbw_{t+1}, \bbLam_t,\bbnu_t)$ associated with variables $\bbw$ and $\bbnu$ are analogously defined to \eqref{eq:proj_grad_def} and \eqref{eq:proj_grad_def2}, respectively. Note descent (respectively, ascent) using projected stochastic gradients in [cf. \eqref{eq:proj_grad_def} - \eqref{eq:proj_grad_def2}] is equivalent to using the projected stochastic saddle point method [cf. \eqref{eq:dict_update} - \eqref{eq:sda_dual_nu_ij}].

To establish convergence of D4L, some conditions are required of the network, loss functions, and stochastic approximation errors which we state below.

%
\vspace{-1mm}
\begin{assumption} \label{as:first} (Network connectivity) The network $\ccalG$ is connected with diameter $D$. The singular values of the incidence matrix $\bbC$ are respectively upper and lower bounded by $\Gamma$ and $\gamma>0$. \end{assumption}
%
\vspace{-2mm}
\begin{assumption} \label{as:2} (Smoothness) 
The Lagrangian has Lipschitz continuous gradients in the primal and dual variables  with constants $L_{ {\bbD}}$, $L_{ {\bbw}}$, $L_{ {\bbLam}}$, and $L_{ {\bbnu}}$. This implies that, e.g.,
\begin{equation} \label{eq:lipschitz_grad}
\| \nabla_{\bbD}\ccalL(\bbD,\bbw, \bbLam,\bbnu)  -  \nabla_{\bbD}\ccalL(\tbD,\bbw, \bbLam,\bbnu) \|  \leq L_{ {\bbD}} \|\bbD - \tbD\|_F \;.
\end{equation} 
Moreover, the projected gradients of the Lagrangian in the primal and dual variables are bounded with block constants  $G_{\bbD}$, $G_{\bbw}$, $G_{\bbLam}$, and $G_{\bbnu}$, which implies that, e.g.,
\begin{equation}\label{eq:grad_bound}
\| \tilde{\nabla}_{\bbD}\ccalL(\bbD,\bbw, \bbLam,\bbnu)\| \leq G_{\bbD} \; .
\end{equation}
\end{assumption}
%
\begin{assumption}\label{as:step_sizes} (Diminishing step-size rules) 
The step-size $\eps_t$ is chosen as $O(1/t)$, i.e. $\eps_t$ satisfies $\\$
 (i) $\sum_{t=1}^\infty \eps_t = \infty$, (non-summability)  $\\$
(ii) $\sum_{t=0}^ \infty \eps_t^2 < \infty $, (square-summability).
\end{assumption}
%
\begin{assumption}\label{as:last} (Stochastic Approximation Error)
The bias of the stochastic gradients of the Lagrangian with respect to each block variable asymptotically converges to null at a rate on the order of the algorithm step-size, which allows us to write, e.g. 
\begin{align}\label{eq:unbiased}
\|  \E{  \bbdelta_{\bbD,t}   \given \ccalF_t  } \| \leq A \eps_t \; ,
\end{align}
where $\bbdelta_{\bbD,t}: =\tilde{\nabla}_\bbD {\ccalL}(\bbD_t, \bbw_t, \bbLam_t, \bbnu_t)  - \tilde{\nabla}_\bbD \hat{\ccalL_t}(\bbD_t, \bbw_t, \bbLam_t, \bbnu_t)  $  denotes the stochastic errors of the Lagrangian with respect to the dictionary $\bbD$, and $\bbdelta_{\bbLam,t}: =\tilde{\nabla}_{\bbLam} {\ccalL}(\bbD_t, \bbw_t, \bbLam_t, \bbnu_t)  - \tilde{\nabla}_{\bbLam} \hat{\ccalL_t}(\bbD_t, \bbw_t, \bbLam_t, \bbnu_t)$ denotes the dual stochastic approximation error with respect to $\bbLam$.

%
%
%
Moreover, let $\ccalF_t$ be a sigma algebra that measures the history of the system up until time $t$. Then, the conditional second moments of the stochastic gradients are bounded by $\sigma^2$ for all times $t$, which for example allows us to write
\begin{align} \label{eq:variance_stochastic_error}
     \E{ \|\tilde{\nabla}_\bbD \hat{\ccalL_t}(\bbD_t, \bbw_t, \bbLam_t, \bbnu_t) \|^2    \given\ccalF_t }
     \leq \sigma^2 .
\end{align}
 \end{assumption}

%
Assumption \ref{as:first} is standard in distributed algorithms and Assumption \ref{as:2} is common in analysis of descent methods, and is guaranteed to hold by making use of gradients which are projected into compact sets $\ccalD^N$, $\ccalW^N$, $\mathfrak{L}^M$, and $\ccalN^M$. Assumption \ref{as:step_sizes} specifies that a diminishing step-size condition for the algorithm and Assumption \ref{as:last} provides conditions on the stochastic approximation errors, both of which are typical in stochastic optimization. They are all satisfied in most cases. 

Observe that the projected stochastic gradients in the updates in \eqref{eq:sp_dict} - \eqref{eq:sp_w} imply that the primal variables themselves are contained in compact sets $\ccalD^N$ and $\ccalW^N$, which allows us to write
\begin{equation}\label{eq:iterate_bound}
\| \bbD \|_F \leq \sqrt{N k}\;, \qquad \|\bbw \| \leq K_{\bbw} \; ,
\end{equation}
for all dictionaries $\bbD\in\ccalD^N$ and model parameters $\bbw\in\ccalW^N$. The compactness of dual sets $\mathfrak{L}$ and $\ccalN$ ensure the primal gradients are bounded [cf.\eqref{eq:grad_bound}], and the respective dual gradients in $\bbLam$ and $\bbnu$ are bounded by constants $G_{\bbLam}=\Gamma \sqrt{N k}$ and $G_{\bbnu}=\Gamma K_{\bbw}$.

Before stating the main theorem, we present a lemma which will be used in its proof, and appears as Proposition 1.2.4 in \cite{Ber99}.

\begin{lemma}\label{lem2}
Let $\{a_t\}$ and $\{b_t\}$ be two nonnegative scalar sequences such that
 $\sum_{t=1}^\infty a_t = \infty$ and
 $\sum_{t=1}^\infty a_t b_t< \infty$.
Then 
\begin{equation} \label{eq:liminf_convergence}
\liminf_{t\rightarrow \infty} b_t = 0.
\end{equation}
 Furthermore, if $| b_{t+1} - b_t |\leq B a_t$ for some constant $B>0$, then 
 \begin{equation} \label{eq:seq_convergence}
\lim_{t\rightarrow \infty} b_t = 0.
\end{equation}
\end{lemma}

With these preliminary results in place, we may state our main result, which says that the proposed algorithm on average asymptotically achieves a first-order stationarity condition of the Lagrangian associated with the optimization problem stated in \eqref{eq:dist_stoch_dict}. 


%
\begin{thm}\label{thm1}
Denote  $(\bbD_t, \bbw_t, \bbLam_t , \bbnu_t )$ as the sequence generated by the block saddle point algorithm in  \eqref{eq:sp_dict}-\eqref{eq:sp_dual_w}. If Assumptions \ref{as:first} - \ref{as:last} hold true, then the first-order stationary condition with respect to the primal variables
\begin{align}\label{eq:thm1}
\lim_{t \rightarrow \infty}&
 	{\mbE} [\| \tilde{\nabla}_{\bbD}\ccalL(\bbD_t, \bbw_t, \bbLam_t, \bbnu_t)  \| ]= 0\; ,\\
\lim_{t \rightarrow \infty}&
	{\mbE} [\| \tilde{\nabla}_{\bbw}\ccalL(\bbD_t, \bbw_t, \bbLam_t, \bbnu_t) \|] = 0 \;
\end{align}
is asymptotically achieved in expectation. Moreover, the asymptotic feasibility condition
\begin{align}\label{eq:thm1_dual}
\lim_{t \rightarrow \infty}&	
{\mbE} [\|\tilde{\nabla}_{\bbLam}\ccalL(\bbD_{t+1}, \bbw_{t+1}, \bbLam_t, \bbnu_t)  \|] = 0\\
\lim_{t \rightarrow \infty}&
{\mbE} [\|\tilde{\nabla}_{\bbnu}\ccalL(\bbD_{t+1}, \bbw_{t+1}, \bbLam_t, \bbnu_t) \|] = 0
\end{align}
is attained in an expected sense.
\end{thm}
%
\begin{myproof} 
The analysis is broken up into distinct components for the primal and dual variables. In the primal variables, we consider the Lagrangian difference of iterates at the next and current time. We expand terms, use properties of the stochastic gradients and function smoothness,  and take conditional expectations on past information to establish a decrement property. We then mirror this analysis in the dual domain. At this point we leverage the step-size rules and apply \eqref{eq:liminf_convergence}. Then we consider the magnitude of block gradient differences which we bound by a term that diminishes with the step-size, which implies \eqref{eq:seq_convergence} holds, yielding an the expected asymptotic convergence to a stationary solution. We subsequently use the shorthand $\nabla_{\bbD} \ccalL(\cdot,\cdot, \cdot,\cdot)_{t+1} : = \nabla_{\bbD} \ccalL(\bbD_{t+1},\bbw_{t+1}, \!\bbLam_t,\bbnu_t)$ and $\nabla_{\bbD} \ccalL(\cdot,\cdot, \cdot,\cdot)_t : = \nabla_{\bbD} \ccalL(\bbD_{t},\bbw_{t}, \!\bbLam_t,\bbnu_t)$, and analogous notation for the other block variables.

Begin by considering the difference of Lagrangians evaluated at the primal variables at the next and current time, and apply Taylor's Theorem to quadratically approximate the former term
\begin{align}\label{eq:lagrangian_dict_diff}
&  \ccalL  (\bbD_{t+1},\bbw_{t+1}, \bbLam_t,\bbnu_t)
 		\! - \! \ccalL  (\bbD_t,\bbw_t, \bbLam_t,\bbnu_t) \nonumber\\
		 &\qquad \leq 
\left[\begin{array}{c}
		\nabla_{\bbD} \ccalL(\cdot,\cdot, \cdot,\cdot)_t \\
		\nabla_{\bbw} \ccalL(\cdot,\cdot, \cdot,\cdot)_t\\
  \end{array} \right] ^T
\left[\begin{array}{c}
		 \bbD_{t+1} \!-\! \bbD_t  \\
		  \bbw_{t+1} \!-\! \bbw_t  \\
  \end{array} \right] \nonumber \\
&\qquad+ {\frac{L_{\bbD}^2}{2}} \|\bbD_{t+1} - \bbD_t\|^2_F + {\frac{L_{\bbw}^2}{2}} \|\bbw_{t+1} - \bbw_t\|^2,
\end{align}
where we have applied the Lipschitz gradient property the Lagrangian to the final two terms as stated in \eqref{eq:lipschitz_grad} to the last term of \eqref{eq:lagrangian_dict_diff}.
The difference of the current and next iterates may be written as $\bbD_{t+1} - \bbD_t = -\eps_t \tilde{\nabla}_\bbD \hat{\ccalL}_t (\bbD_t,\bbw_t, \bbLam_t,\bbnu_t)$ and $\bbw_{t+1} - \bbw_t = -\eps_t \tilde{\nabla}_\bbw \hat{\ccalL}_t (\bbD_t,\bbw_t, \bbLam_t,\bbnu_t)$, which we substitute into the right hand side of \eqref{eq:lagrangian_dict_diff}, yielding
\begin{align} \label{eq:stoch_grad_subs_dict}
&  \ccalL  (\bbD_{t+1},\bbw_{t+1}, \bbLam_t,\bbnu_t)
 		\! - \! \ccalL  (\bbD_t,\bbw_t, \bbLam_t,\bbnu_t) \\
		 & \leq \!-\eps_t\!\!
\left[\begin{array}{c}
	\!\!	\nabla_{\bbD} \ccalL(\cdot,\cdot, \cdot,\cdot)_t \\
	\!\!	\nabla_{\bbw} \ccalL(\cdot,\cdot, \cdot,\cdot)_t\\
  \end{array} \right] ^T\!\!
\left[\begin{array}{c}
	\!\!	  \tilde{\nabla}_\bbD \hat{\ccalL}_t (\cdot,\cdot, \cdot,\cdot)_t\\
	\!\!	   \tilde{\nabla}_\bbw \hat{\ccalL}_t(\cdot,\cdot, \cdot,\cdot)_t\\
  \end{array}\!\! \right] \nonumber \\
&\quad+\!   \frac{\eps_t^2}{2} \Big(  {L_{\bbD}^2} \|\tilde{\nabla}_\bbD \hat{\ccalL}_t (\cdot,\cdot, \cdot,\cdot)_t\|^2_F \! 
+  {L_{\bbw}^2} \|\tilde{\nabla}_\bbw \hat{\ccalL}_t (\cdot,\cdot, \cdot,\cdot)_t\|^2 \Big), \nonumber
\end{align}
Take the expectation of \eqref{eq:stoch_grad_subs_dict} conditional on the filtration $\ccalF_t$, apply the finite conditional variance condition [cf. \eqref{eq:variance_stochastic_error}] stated in Assumption and substitute in the definition of the projected stochastic gradient error of the Lagrangian with respect to $\bbD$ to write
\begin{align} \label{eq:conditional_mean_dict1}
&{\mbE}[   \ccalL  (\bbD_{t+1},\bbw_{t+1}, \bbLam_t,\bbnu_t)
 		\! - \! \ccalL  (\bbD_t,\bbw_t, \bbLam_t,\bbnu_t)\mid \ccalF_t] \\
		 & \leq \!-\eps_t\!\!
\left[\begin{array}{c}
	\!\!	\nabla_{\bbD} \ccalL(\cdot,\cdot, \cdot,\cdot)_t \\
	\!\!	\nabla_{\bbw} \ccalL(\cdot,\cdot, \cdot,\cdot)_t\\
  \end{array} \right] ^T\!\!
\left[\begin{array}{c}
	\!\!	  \tilde{\nabla}_\bbD {\ccalL} (\cdot,\cdot, \cdot,\cdot)_t - \bbdelta_{\bbD,t}\\
	\!\!	   \tilde{\nabla}_\bbw {\ccalL}(\cdot,\cdot, \cdot,\cdot)_t- \bbdelta_{\bbw,t}\\
  \end{array}\!\! \right] \nonumber \\
&\quad
		 + \!\frac{\eps_t^2 \sigma^2}{2}\Big(L_{\bbD}  \!+\! L_{\bbw} \Big) , \nonumber
\end{align}
Now we apply the fact that the direction defined by the projected gradient method is a descent direction, which appears, for instance, as \cite{Birgin00nonmonotonespectral}, Lemma 2.1(i), to the first term of \eqref{eq:conditional_mean_dict1}, which allows us to  write
\begin{align} \label{eq:conditional_mean_dict2}
{\mbE}[   \ccalL & (\bbD_{t+1},\bbw_{t+1}, \bbLam_t,\bbnu_t)
 		\! - \! \ccalL  (\bbD_t,\bbw_t, \bbLam_t,\bbnu_t)\mid \ccalF_t] \nonumber \\
		  &\leq \!-\eps_t
 \left\|\begin{array}{c}
	\!\!	\tilde{\nabla}_{\bbD} \ccalL_t(\bbD_t,\bbw_t, \bbLam_t,\bbnu_t ) \!\!\\
	\!\!	\tilde{\nabla}_{\bbw} \ccalL_t(\bbD_t,\bbw_t, \bbLam_t,\bbnu_t ) \!\!\\
  \end{array} \right\|^2\!\!\!\!\!\!  \nonumber \\ &\quad
	+ \eps_t^2 \Big( A (G_{\bbD} + G_{\bbW})	 + \!\frac{ \sigma^2}{2}\big(L_{\bbD}  \!+\! L_{\bbw} \big) \Big) , 
\end{align}
where we used the Cauchy-Schwartz inequality, the bias condition in \eqref{eq:unbiased}, and the bound on the partial gradients of the Lagrangian with respect to $\bbD$ and $\bbw$ as stated in \eqref{eq:grad_bound} of Assumption \ref{as:2} to the second term inside the brackets on the right hand side of \eqref{eq:conditional_mean_dict1}. 

Set this analysis aside for now and consider Taylor expansion around Lagrangian evaluated at the dual iterates at the next and current time, which since the Lagrangian is a linear function of its multipliers, allows us to write
\begin{align}\label{eq:lagrangian_dual_diff}
& \ccalL  (\bbD_{t+1},\bbw_{t+1}, \bbLam_{t+1},\bbnu_{t+1})
 		\! - \!  \ccalL  (\bbD_{t+1},\bbw_{t+1}, \bbLam_t,\bbnu_t)  \nonumber\\
		 &\qquad = 
\left[\begin{array}{c}
		\nabla_{\bbLam} \ccalL (\cdot,\cdot, \cdot,\cdot)_{t+1} \\
		\nabla_{\bbnu} \ccalL (\cdot,\cdot, \cdot,\cdot)_{t+1}\\
  \end{array} \right] ^T
\left[\begin{array}{c}
		 \bbLam_{t+1} \!-\! \bbLam_t  \\
		  \bbnu_{t+1} \!-\! \bbnu_t  \\
  \end{array} \right] 
\end{align}
The difference of the current and next iterates may be written as $\bbLam_{t+1} - \bbLam_t = \eps_t \tilde{\nabla}_{\bbLam} \hat{\ccalL} (\bbD_{t+1},\bbw_{t+1}, \bbLam_t,\bbnu_t)$ and $\bbnu_{t+1} - \bbnu_t = \eps_t \tilde{\nabla}_{\bbnu} \hat{\ccalL} (\bbD_{t+1},\bbw_{t+1}, \bbLam_t,\bbnu_t)$, which we substitute into the right hand side of \eqref{eq:lagrangian_dual_diff}, yielding
\begin{align} \label{eq:stoch_grad_subs_dual}
  \ccalL  (\bbD_{t+1},\!\bbw_{t+1}, &\bbLam_{t+1},\!\bbnu_{t+1})
 		\! - \! \ccalL  (\bbD_{t+1},\!\bbw_{t+1}, \bbLam_t,\!\bbnu_t) \\
		 & = \! \eps_t\!
\left[\!\!\!\begin{array}{c}
		\!\nabla_{\bbLam} \ccalL(\cdot,\cdot, \cdot,\cdot)_{t+1} \!\\
		\!\nabla_{\bbnu} \ccalL(\cdot,\cdot, \cdot,\cdot)_{t+1}\!\\
  \end{array}\!\!\! \right] ^T
\!\!\!\left[\!\!\!\begin{array}{c}
		  \tilde{\nabla}_{\bbLam} \hat{\ccalL}_t (\cdot,\cdot, \cdot,\cdot)_{t+1}\!  \\
		   \tilde{\nabla}_{\bbnu} \hat{\ccalL}_t (\cdot,\cdot, \cdot,\cdot)_{t+1}\!  \\
  \end{array}\!\!\! \right] \nonumber
\end{align}
Substitute the definition of the dual stochastic gradient errors $\bbdelta_{\bbLam,t }$, $\bbdelta_{\bbnu,t }$ in Assumption \ref{as:last} into the right hand side of \eqref{eq:stoch_grad_subs_dual} to obtain
\begin{align} \label{eq:stoch_dual_grad_errors}
&  \ccalL  (\bbD_{t+1},\!\bbw_{t+1}, \bbLam_{t+1},\!\bbnu_{t+1})
 		\! - \! \ccalL  (\bbD_{t+1},\!\bbw_{t+1}, \bbLam_t,\!\bbnu_t) \\
		 & \qquad=\! \eps_t\!
 \left[\begin{array}{c}
	\!\!	\nabla_{\bbLam} \ccalL(\cdot,\cdot, \cdot,\cdot)_{t+1} \\
	\!\!	\nabla_{\bbnu} \ccalL(\cdot,\cdot, \cdot,\cdot)_{t+1}\\
  \end{array} \right] ^T\!\!
\left[\begin{array}{c}
	\!\!	  \tilde{\nabla}_{\bbLam} {\ccalL} (\cdot,\cdot, \cdot,\cdot)_{t+1} - \bbdelta_{\bbLam,t}\\
	\!\!	   \tilde{\nabla}_{\bbnu} {\ccalL}(\cdot,\cdot, \cdot,\cdot)_{t+1} - \bbdelta_{\bbnu,t}\\
  \end{array}\!\! \right] \nonumber
\end{align}
Applying the fact that the direction defined by the projected gradient method is an ascent direction,  [\cite{Birgin00nonmonotonespectral}, Lemma 2.1(i)], and computing the expectation conditional on the algorithm history $\ccalF_t$ up to time $t$, we may write
\begin{align}  \label{eq:conditional_mean_dual}
{\mbE}[ \ccalL & (\bbD_{t+1},\!\bbw_{t+1}, \bbLam_{t+1},\!\bbnu_{t+1})
 		\! - \! \ccalL  (\bbD_{t+1},\!\bbw_{t+1}, \bbLam_t,\!\bbnu_t)
		 \!\mid \!\ccalF_t] \nonumber\\
		&\!\geq \eps_t \left\|\!\begin{array}{c}
		\! \tilde{\nabla}_{\bbLam} \ccalL_t(\cdot,\cdot, \cdot,\cdot)_{t+1} \!\\
		\! \tilde{\nabla}_{\bbnu} \ccalL_t(\cdot,\cdot, \cdot,\cdot)_{t+1}\!\\
  \end{array}\! \right \| ^2 - \eps_t^2 A (G_{\bbLam} + G_{\bbnu}) \; ,
\end{align}
 where we have also applied the bias condition of the dual stochastic errors of the Lagrangian [cf. \eqref{eq:unbiased}]  in Assumption \ref{as:last} to the last term. 
 We establish a martingale relationship of the projected primal and dual stochastic gradients by summing the relation in \eqref{eq:conditional_mean_dict1} and with the negative of \eqref{eq:conditional_mean_dual}, which yields
\begin{align}\label{eq:primal_dual_seq}
{\mbE}[  \ccalL  (&\bbD_{t+1},\bbw_{t+1}, \bbLam_{t+1},\bbnu_{t+1})
- \ccalL  (\bbD_{t},\bbw_t,\! \bbLam_{t},\bbnu_t\!)
		 \!\mid \!\ccalF_t] \\
		&\leq  -\eps_t \Big( \left\|\begin{array}{c}
	\!\!	\tilde{\nabla}_{\bbD} \ccalL(\bbD_t,\bbw_t, \bbLam_t,\bbnu_t ) \\
	\!\!	\tilde{\nabla}_{\bbw} \ccalL(\bbD_t,\bbw_t, \bbLam_t,\bbnu_t )\\
  \end{array} \right\|^2 \nonumber \\
&\quad + \left\|\!\begin{array}{c}
		\!\tilde{\nabla}_{\bbLam} \ccalL(\bbD_{t+1},\bbw_{t+1},\! \bbLam_t,\bbnu_t)\!\\
		\!\tilde{\nabla}_{\bbnu} \ccalL (\bbD_{t+1},\bbw_{t+1},\! \bbLam_t,\bbnu_t)\!\\
  \end{array}\! \right \| ^2 \Big)\! \nonumber \\
  &\quad +\! \eps_t^2 \Big(A\big(G_{\bbD}  \!+\! G_{\bbw} + G_{\bbLam} + G_{\bbnu}\big) \! 
		 + \!\frac{\sigma^2}{2}\big(L_{\bbD}  \!+\! L_{\bbw} \big)\!\Big) \; .\nonumber
\end{align}
Observe that the left hand side of \eqref{eq:primal_dual_seq} is telescopic, and hence if we sum this relation over all $t$ we obtain a finite quantity in expectation. By applying the step-size rules stated in Assumption \ref{as:step_sizes} with the fact that $\ccalL$ is lower-bounded since the primal and dual domains are compact, the following holds in expectation 
\begin{align} \label{eq:primal_dual_summable}
 \sum_{t=1}^\infty &
 \eps_t {\mbE}\Big[ \left\|\begin{array}{c}
	\!\!	\tilde{\nabla}_{\bbD} \ccalL_t(\bbD_t,\bbw_t, \bbLam_t,\bbnu_t ) \\
	\!\!	\tilde{\nabla}_{\bbw} \ccalL_t(\bbD_t,\bbw_t, \bbLam_t,\bbnu_t )\\
  \end{array} \right\|^2 \nonumber \\
&\quad + \left\|\!\begin{array}{c}
		\!\tilde{\nabla}_{\bbLam} \ccalL_t(\bbD_{t+1},\bbw_{t+1}, \!\bbLam_t,\bbnu_t) \!\\
		\!\tilde{\nabla}_{\bbnu} \ccalL_t(\bbD_{t+1},\bbw_{t+1},\! \bbLam_t,\bbnu_t)\!\\
  \end{array}\! \right \| ^2 \Big] < \infty \; , 
\end{align}
which as a consequence of \eqref{eq:liminf_convergence} as stated in Lemma \ref{lem2}, implies that the expected limit infinum of the sequence converges to null, 

\begin{align} \label{eq:dict_liminf_convergence}
 \liminf_{t\rightarrow\infty}& \ {\mbE}\Big[ \left\|\begin{array}{c}
	\!\!	\tilde{\nabla}_{\bbD} \ccalL_t(\bbD_t,\bbw_t, \bbLam_t,\bbnu_t ) \\
	\!\!	\tilde{\nabla}_{\bbw} \ccalL_t(\bbD_t,\bbw_t, \bbLam_t,\bbnu_t )\\
  \end{array} \right\|^2 \nonumber \\
&\quad+ \left\|\!\begin{array}{c}
		\!\tilde{\nabla}_{\bbLam} \ccalL_t(\bbD_{t+1},\bbw_{t+1}, \!\bbLam_t,\bbnu_t) \!\\
		\!\tilde{\nabla}_{\bbnu} \ccalL_t(\bbD_{t+1},\bbw_{t+1},\! \bbLam_t,\bbnu_t)\!\\
  \end{array}\! \right \| ^2 \Big] = 0 
\end{align}
Using the convergence in \eqref{eq:dict_liminf_convergence}, we may establish the whole sequence of partial gradients of the Lagrangian converge. Since the logic is equivalent in each block variable, it is enough to consider just the primal-dual pair $(\bbD, \bbLam)$.  Consider the expected absolute difference of the Lagrangian gradients evaluated at the next and current iterate, and apply Jensen's inequality to write
\begin{align} \label{eq:dict_abs_grad_diff}
\Big|\|{\mbE}&[ \tilde{\nabla}_{\bbD} \ccalL(\cdot,\cdot, \cdot,\cdot)_{t+1} ]\|^2_F
		 \!- \!\|{\mbE} [\tilde{\nabla}_{\bbD} \ccalL(\cdot,\cdot, \cdot,\cdot)_{t} ]\|^2_F\Big|\nonumber\\
		&\leq {\mbE}[\| \tilde{\nabla}_{\bbD} \ccalL(\cdot,\cdot, \cdot,\cdot)_{t+1} 
		 \!+ \! \tilde{\nabla}_{\bbD} \ccalL(\cdot,\cdot, \cdot,\cdot)_{t} \|_F \nonumber \\
&\quad \times		 \|\tilde{\nabla}_{\bbD} \ccalL(\cdot,\cdot, \cdot,\cdot)_{t+1} 
		 \!-\!\tilde{\nabla}_{\bbD} \ccalL(\cdot,\cdot, \cdot,\cdot)_{t} \|_F	] \; .	 
\end{align}
Apply the non-expansive property of the projection operator, the bound on the primal gradient in \eqref{eq:grad_bound}, Lipschitz continuity to right hand side of \eqref{eq:dict_abs_grad_diff} to express this gradient difference in terms of the difference between the next and current iterate as
\begin{align} \label{eq:dict_grad_lipschitz}
 {\mbE}[\| & \tilde{\nabla}_{\bbD} \ccalL(\cdot,\cdot, \cdot,\cdot)_{t+1} 
		 +  \tilde{\nabla}_{\bbD} \ccalL(\cdot,\cdot, \cdot,\cdot)_{t} \|_F \nonumber \\
		&\times \|\tilde{\nabla}_{\bbD} \ccalL(\cdot,\cdot, \cdot,\cdot)_{t+1} 
		 -\tilde{\nabla}_{\bbD} \ccalL(\cdot,\cdot, \cdot,\cdot)_{t} \|_F	] \nonumber\\
		 &\leq  
2 G_{\bbD} \|{\nabla}_{\bbD} \ccalL(\cdot,\cdot, \cdot,\cdot)_{t+1} 
		 -{\nabla}_{\bbD} \ccalL(\cdot,\cdot, \cdot,\cdot)_{t} \|_F	] \nonumber\\
		 &\leq 2 G_{\bbD} L_{\bbD}
		 {\mbE}[\|\bbD_{t+1} - \bbD_t\|] \; .
\end{align}
Substitute $\bbD_{t+1} - \bbD_t=-\eps_t \tilde{\nabla}_\bbD \hat{\ccalL}_t (\bbD_t,\bbw_t, \bbLam_t,\bbnu_t)$ into the right hand side of \eqref{eq:dict_grad_lipschitz} and apply with the bound on the second conditional moment of the stochastic gradient stated in \eqref{eq:variance_stochastic_error} of Assumption \ref{as:last} to write
\begin{align} \label{eq:dict_grad_lipschitz}
2 G_{\bbD} L_{\bbD} \eps_t{\mbE}[\|\tilde{\nabla}_\bbD \hat{\ccalL}_t (\bbD_t,\bbw_t, \bbLam_t,\bbnu_t)\| ]
	\leq 2 G_{\bbD} L_{\bbD} \sigma \eps_t \; .
\end{align}
With \eqref{eq:dict_grad_lipschitz}, the second condition of Lemma \ref{lem2} is satisfied, whereby we may conclude the gradient sequence converges in expectation
\begin{equation}\label{eq:primal_grad_convergence1}
\lim_{t\rightarrow \infty} {\mbE}[ \|\tilde{\nabla}_{\bbD}\ccalL(\bbD_t,\bbw_t, \bbLam_t,\bbnu_t) \| ] =0 \; .
\end{equation}
%
%
Since the other primal block sequence $\bbw_t$ is updated in an analogous manner to that of $\bbD$, the analysis with the same logic, implying that a first order stationary condition of the Lagrangian is achieved asymptotically in expectation, i.e. $ {\mbE}[\|\tilde{\nabla}_{\bbw}\ccalL(\bbD_t,\bbw_t, \bbLam_t,\bbnu_t)\| ]\rightarrow 0$. 

We next establish that the whole dual gradient sequence with respect to $\bbLam$ is converging in magnitude to null. We use the shorthand notation $\nabla_{\bbLam} \ccalL(\cdot,\cdot, \cdot,\cdot)_{t+1} : = \nabla_{\bbLam} \ccalL(\bbD_{t+1},\bbw_{t+1}, \!\bbLam_{t+1},\bbnu_{t+1})$ and $\nabla_{\bbLam} \ccalL(\cdot,\cdot, \cdot,\cdot)_t : = \nabla_{\bbLam} \ccalL(\bbD_{t+1},\bbw_{t+1}, \!\bbLam_t,\bbnu_t)$. Continue by considering the expected absolute difference of the Lagrangian gradients evaluated at the next and current dual iterate, and applying Jensen's inequality to write
\begin{align} \label{eq:dual_abs_grad_diff}
&\Big|\|{\mbE}[ \tilde{\nabla}_{\bbLam} \ccalL(\cdot,\cdot, \cdot,\cdot)_{t+1} ]\|^2_F
		 - \|{\mbE} [\tilde{\nabla}_{\bbLam} \ccalL(\cdot,\cdot, \cdot,\cdot)_t ]\|^2_F\Big|\nonumber\\
		&\quad \leq {\mbE}[\| \tilde{\nabla}_{\bbLam} \ccalL(\cdot,\cdot, \cdot,\cdot)_{t+1} 
		 +  \tilde{\nabla}_{\bbLam} \ccalL(\cdot,\cdot, \cdot,\cdot)_t \|_F \nonumber \\
		&\qquad\times \|\tilde{\nabla}_{\bbLam} \ccalL(\cdot,\cdot, \cdot,\cdot)_{t+1} 
		 -\tilde{\nabla}_{\bbLam} \ccalL(\cdot,\cdot, \cdot,\cdot)_t \|_F	]	 \nonumber \\
		 &\quad= {\mbE}[\| \bbC_{\bbD} \bbD_{t+1} 
		 + \bbC_{\bbD} \bbD_{t} \|_F 
		 \|\bbC_{\bbD} \bbD_{t+1} 
		\! -\! \bbC_{\bbD} \bbD_{t} \|_F	]			 
\end{align}
where the last equality in \eqref{eq:dual_abs_grad_diff} follows from the computation of the dual gradient of the Lagrangian in \eqref{eq:dual_grad_lam_appendix}. Now apply the triangle inequality and the compactness of the set $\ccalD$ to express the right hand side of \eqref{eq:dual_abs_grad_diff} in terms of the difference between the next and current iterate, yielding
\begin{align} \label{eq:dict_grad_primal_subs}
 {\mbE}&[\| \bbC_{\bbD} \bbD_{t+1} 
		 + \bbC_{\bbD} \bbD_{t} \|_F 
		 \|\bbC_{\bbD} \bbD_{t+1} 
		\! -\! \bbC_{\bbD} \bbD_{t} \|_F	] \nonumber \\
		 &\leq   2 \Gamma \sqrt{Nk} {\mbE}[\|\bbD_{t+1} \!- \! \bbD_t\|] \; ,  \\
		 &=   2 \Gamma \sqrt{Nk} \eps_t {\mbE}[\|\tilde{\nabla}_\bbD \hat{\ccalL}_t (\bbD_t,\bbw_t, \bbLam_t,\bbnu_t)\|]  
		 \leq   2 \Gamma \sqrt{Nk}  \sigma\eps_t \; , \nonumber
\end{align}
where the second equality comes from the substitution  $\bbD_{t+1} - \bbD_t=-\eps_t\nabla_\bbD \hat{\ccalL}(\bbD_t,\bbw_t, \bbLam_t,\bbnu_t)$, and the last inequality comes from applying the bound in \eqref{eq:variance_stochastic_error}. As in the analysis of the primal gradient sequence, we may now apply \eqref{eq:seq_convergence} in Lemma \ref{lem2}, which implies that the expected projected dual gradient sequence converges to null in magnitude, i.e.
\begin{equation}\label{eq:dual_grad_convergence1}
\lim_{t\rightarrow \infty}{\mbE} [ \|\tilde{\nabla}_{\bbLam}\ccalL(\bbD_{t+1},\bbw_{t+1}, \bbLam_t,\bbnu_t) \| ]=0.
\end{equation}
%
%
By noting that the analysis for $\bbLam$ is analogous to that of the other dual variable $\bbnu$, we may also conclude $ {\mbE} [\|\tilde{\nabla}_{\bbnu}\ccalL(\bbD_{t+1},\bbw_{t+1}, \bbLam_t,\bbnu_t)\|] \rightarrow 0$. \end{myproof} 

Theorem \ref{thm1} guarantees that the block saddle point method as stated in \eqref{eq:sp_dict} - \eqref{eq:sp_dual_w} solves the problem of learning a dictionary and discriminative model over that dictionary representation of the feature space in a decentralized online manner. In particular, the algorithm asymptotically converges to a KKT point of the problem \eqref{eq:dist_stoch_dict} in expectation. This implies that the primal variables converge to a local minimum of the objective, and the dual variables asymptotically enforce feasibility, i.e. the network agreement constraints are satisfied in expectation. We next turn to the practical consequences of this theorem by studying the algorithm performance on a canonical computer vision task.

\section{Simulations}\label{sec:experiments}
%
\begin{figure}
	\centering \vspace{2mm}
	\includegraphics[width=0.38\linewidth]{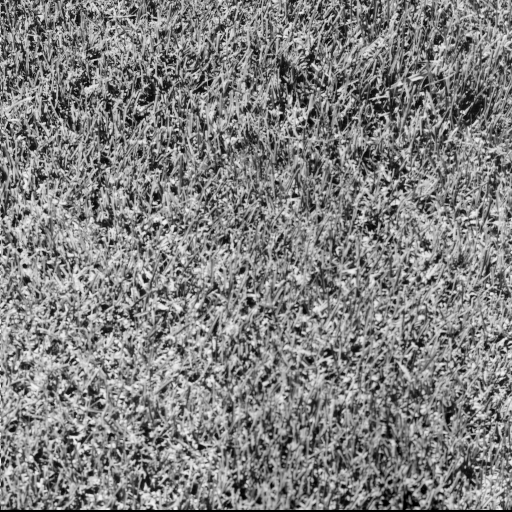}
	\hspace{5mm}
	\includegraphics[width=0.38\linewidth]{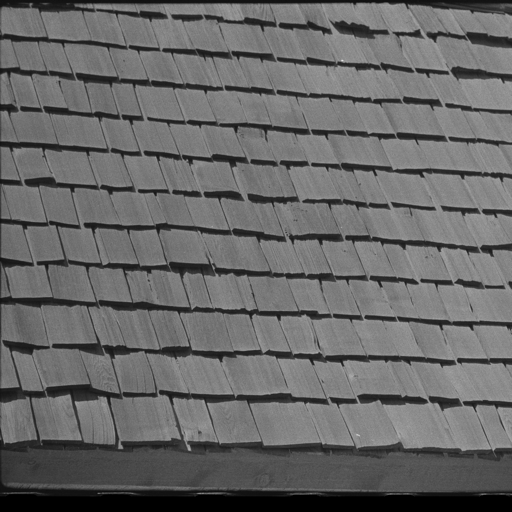}
	\caption{Sample images from the Brodatz texture database.}
	\label{fig:exampledata}
\end{figure}

%
\begin{figure}
	\centering
	\includegraphics[width=0.20\textwidth]{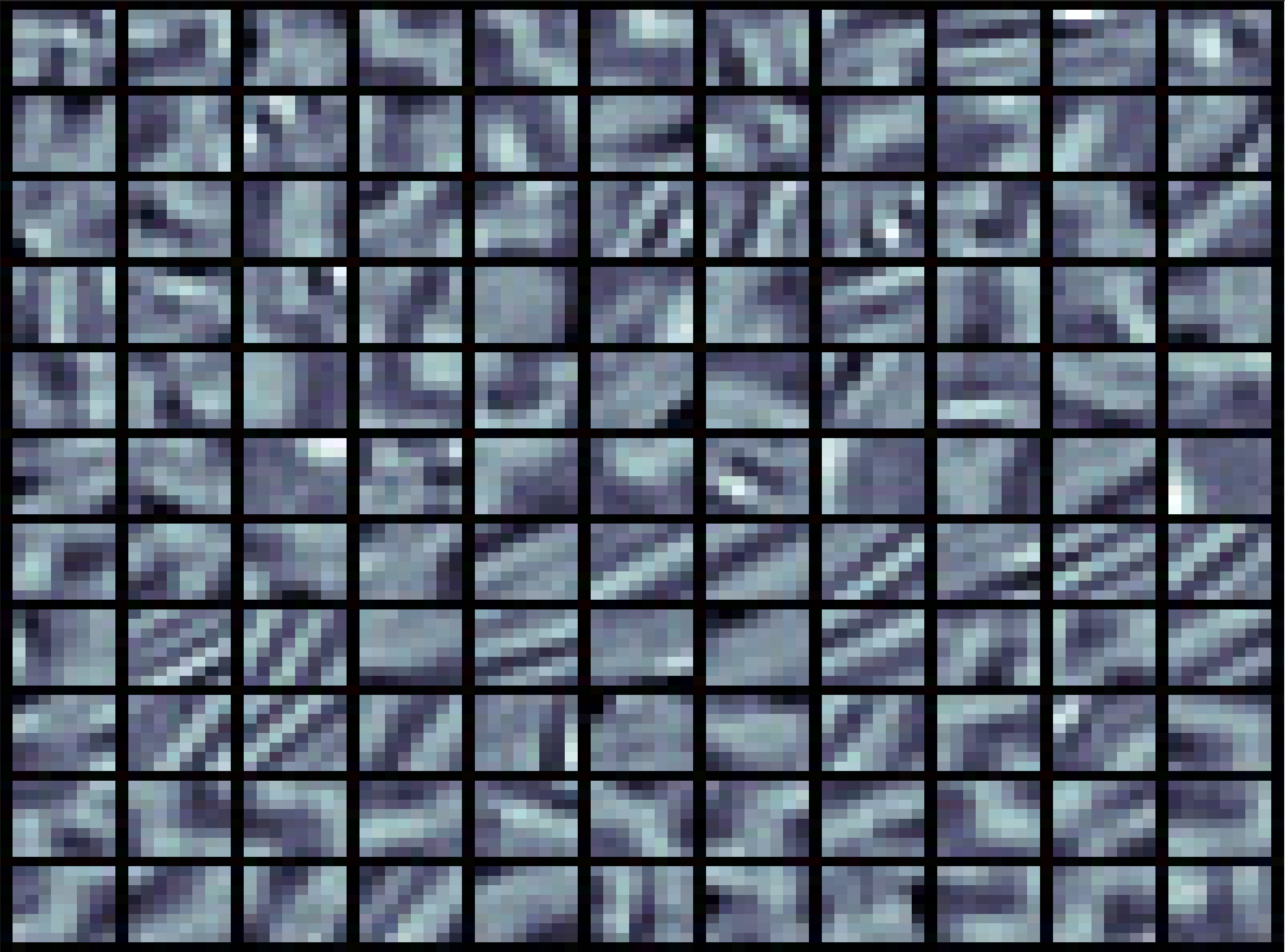}
	\hspace{0.5cm}
	\includegraphics[width=0.20\textwidth]{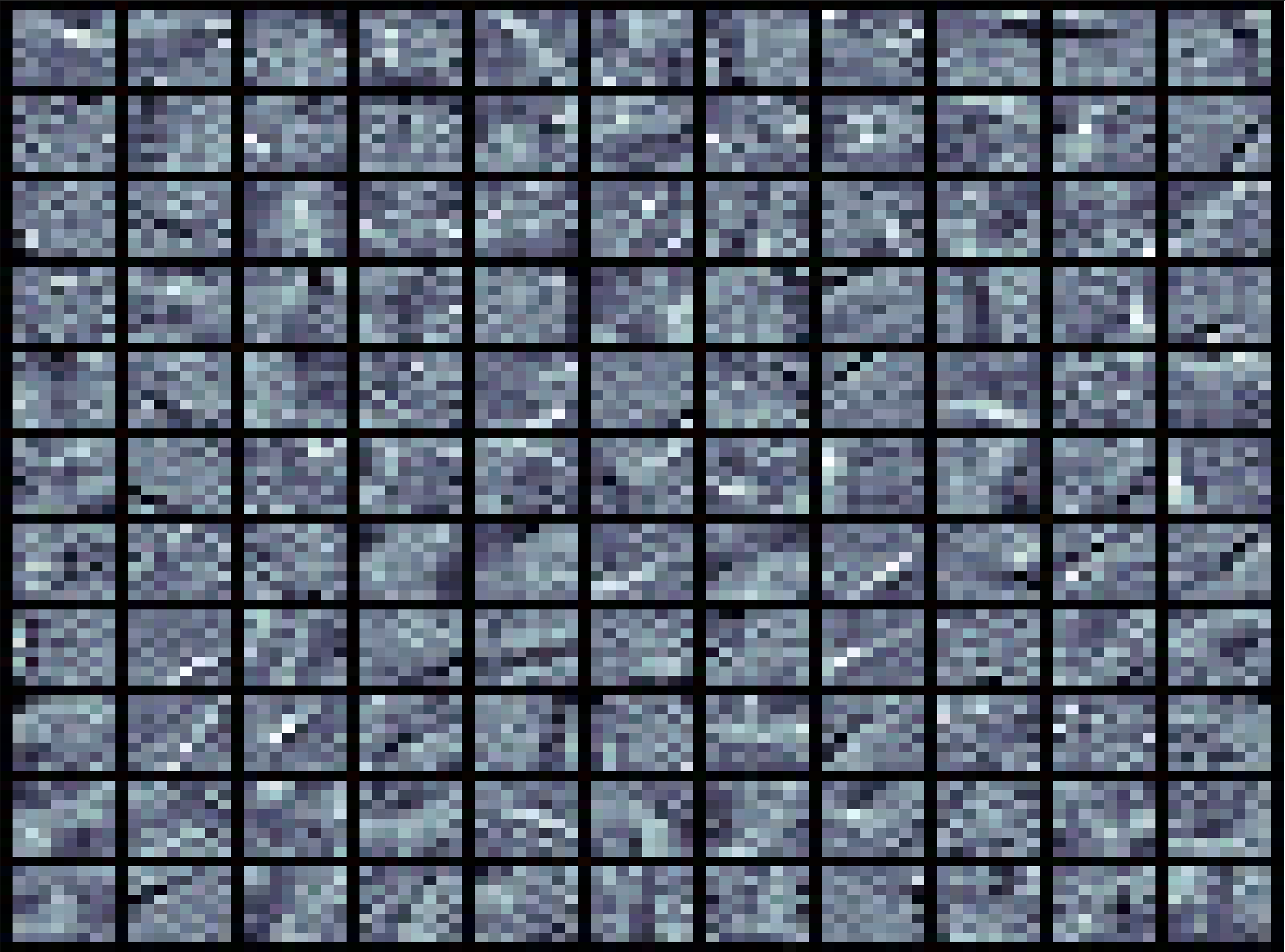}
	\caption{Initialized (left) and final (right) dictionary for $8$-by-$8$ grayscale patches.  These dictionaries were computed using the centralized ($N=1$) algorithm with step-size $\epsilon=0.25$.}
	\label{fig:initdictionary}\vspace{-5mm}
\end{figure}

Consider the task of visual pattern recognition in large scale image databases. Because the sample size is assumed to be very large, learning over the data all at once is impractical. Instead, we consider processing images a few at a time. Moreover, image processing is computationally demanding domain in which learning at a centralized location may be too slow. By leveraging a network of interconnected computing nodes and using Algorithm \ref{alg:d4l}, we may effectively accelerate the rate at which such large-scale pattern recognition tasks may be solved. 
%
\begin{figure*}
\begin{subfigure}{.5\columnwidth}
\includegraphics[width=\linewidth, height = 0.5\linewidth]
                {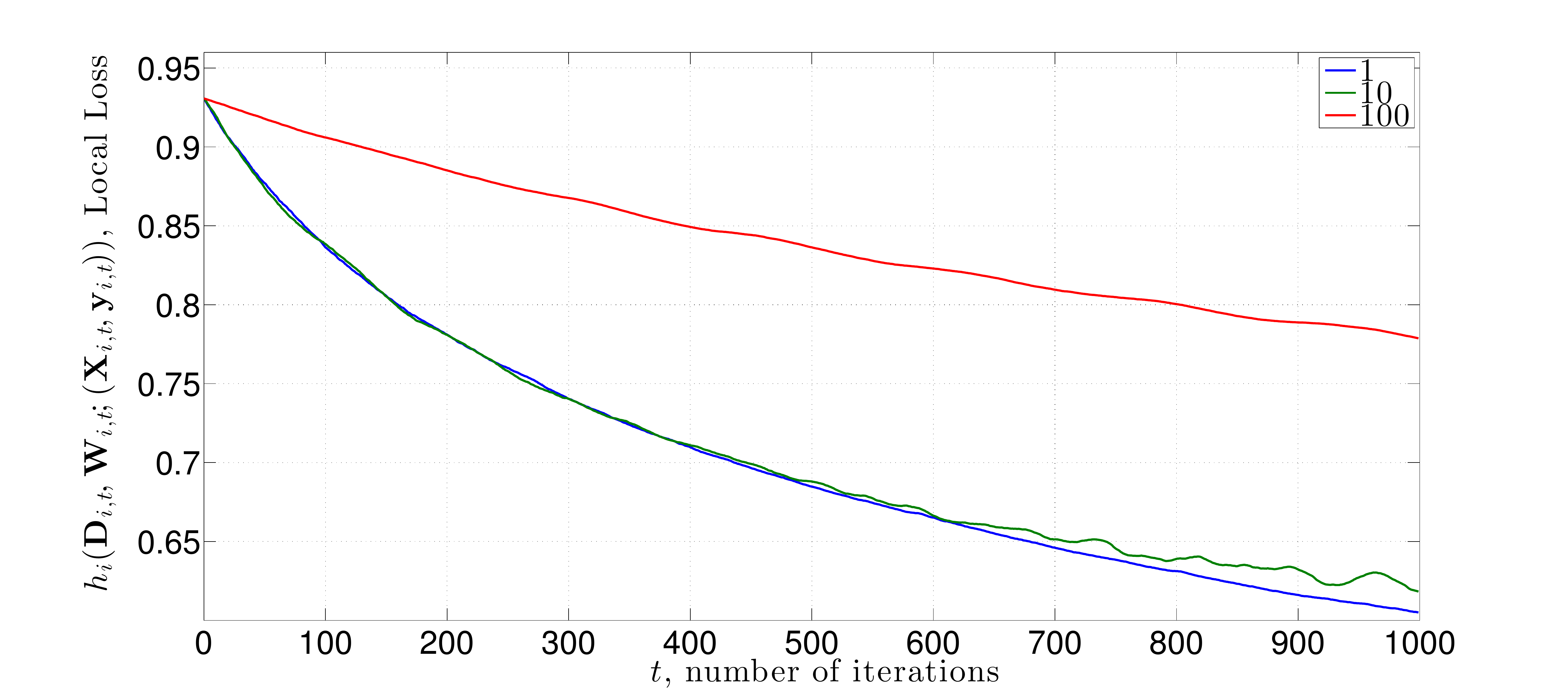}\vspace{-2mm}
\caption{Local loss versus iteration $t$}
\label{subfiga_vary_n}
\end{subfigure}
\begin{subfigure}{.5\columnwidth}
\includegraphics[width=\linewidth,height = 0.5\linewidth]
                {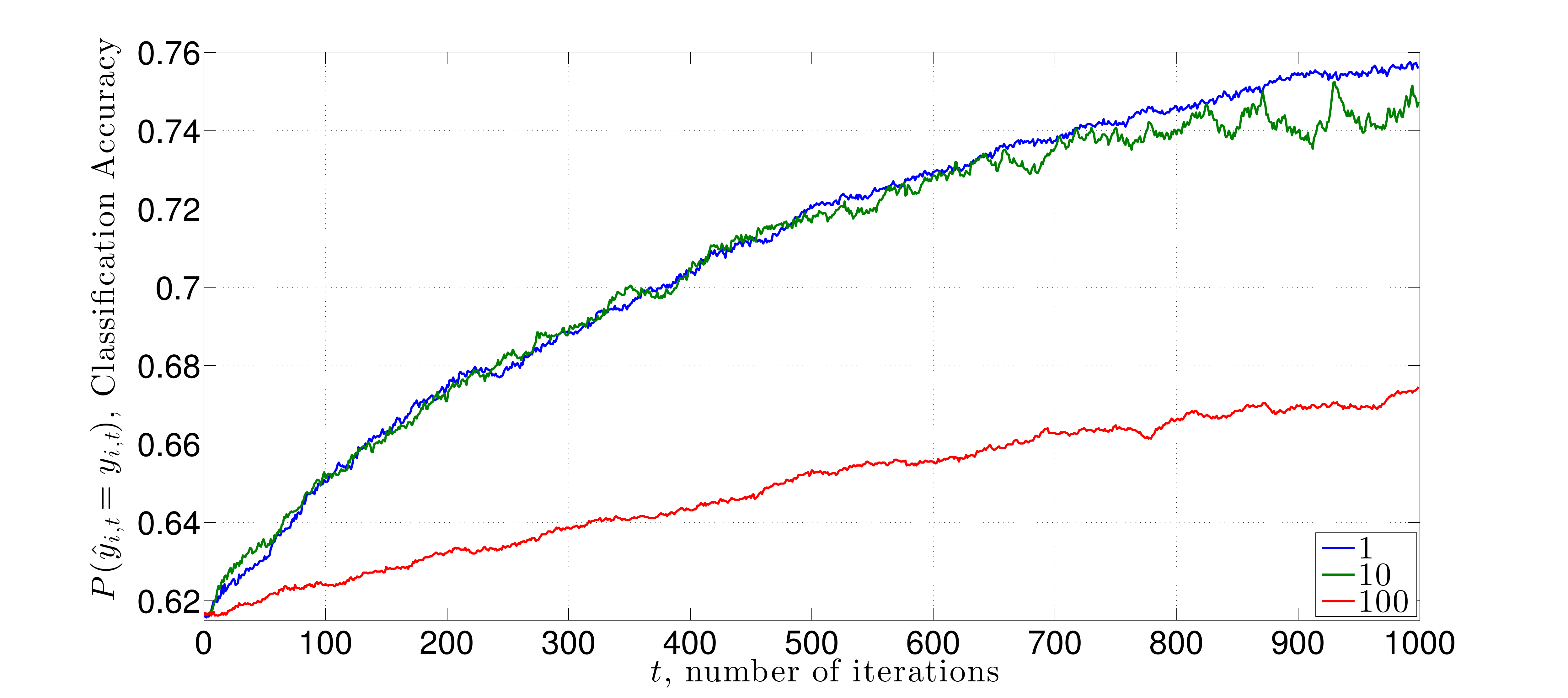}\vspace{-2mm}
\caption{Classification accuracy vs. iteration $t$}
\label{subfigb_vary_n}
\end{subfigure}
\begin{subfigure}{.5\columnwidth}
\includegraphics[width=\linewidth,height = 0.5\linewidth]
                {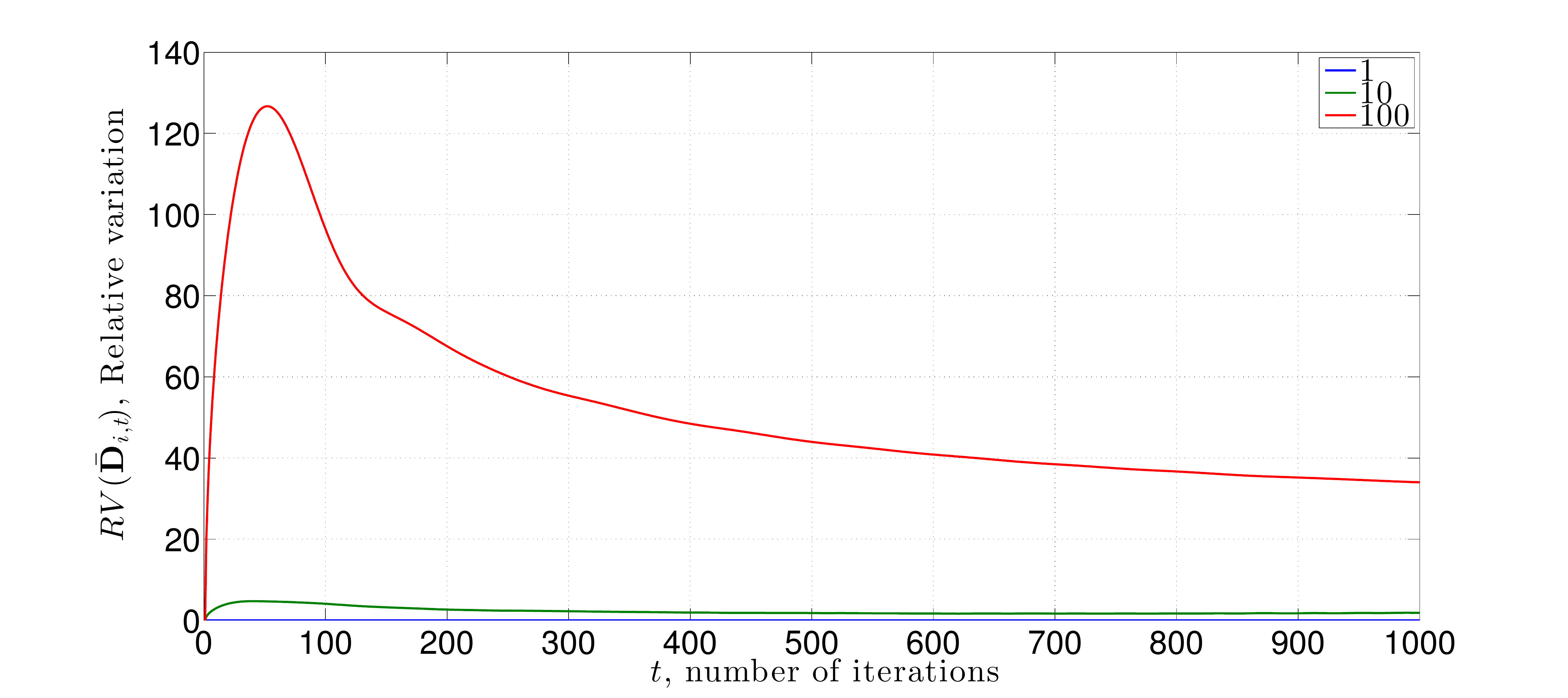}\vspace{-2mm}
\caption{Relative variation $\RV(\bar{\bbD}_{i,t})$ vs. iteration $t$}
\label{subfigc_vary_n}
\end{subfigure}%
\begin{subfigure}{.5\columnwidth}
\includegraphics[width=\linewidth,height = 0.5\linewidth]
                {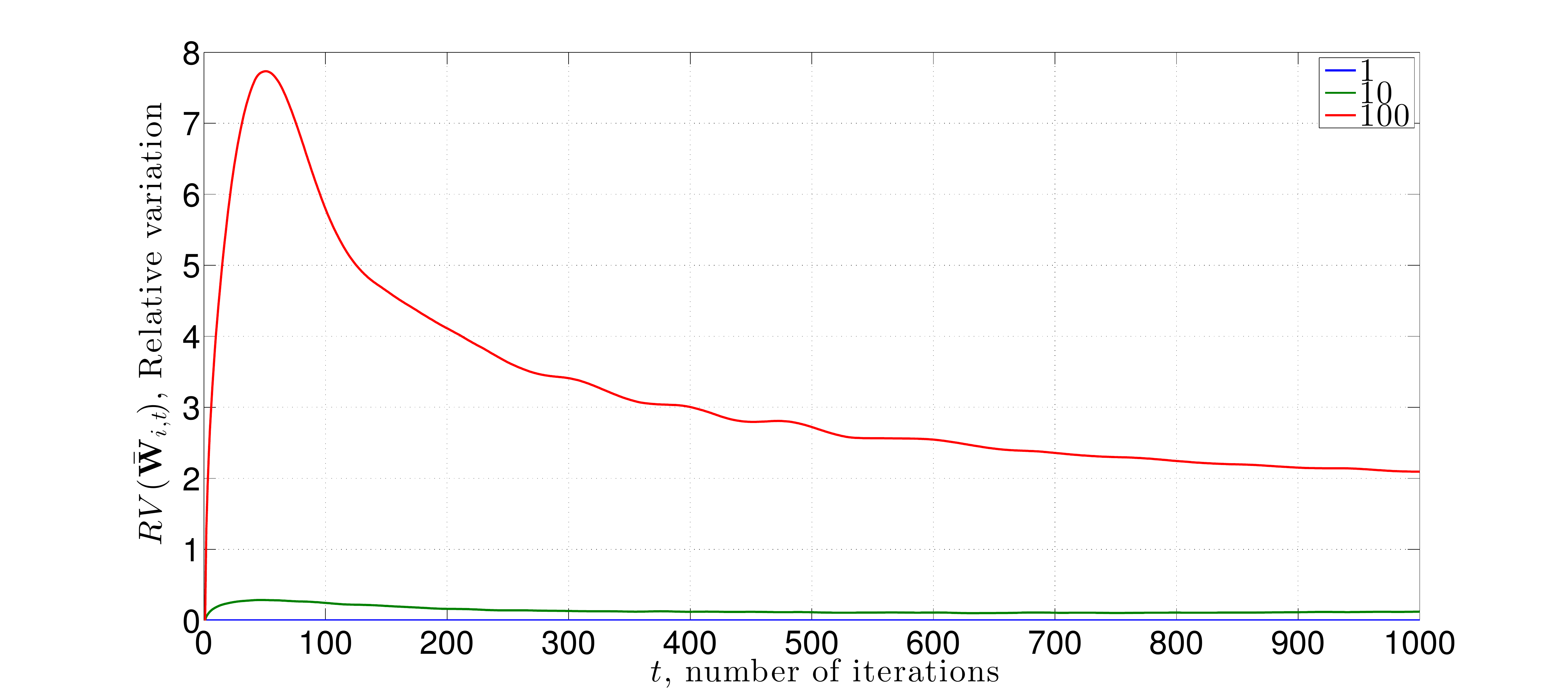}\vspace{-2mm}
\caption{Relative variation $\RV(\bar{\bbW}_{i,t})$ vs. iteration $t$}
\label{subfigd_vary_n}
\end{subfigure}%
\caption{Learning achieved by an arbitrary agent in networks of size $N=1$ (centralized), $N=10$, and $N=100$ with nodes randomly connected with prob. $\rho=0.2$. \ref{subfiga_vary_n}-\ref{subfigb_vary_n} show $h_{i} (\bbD_{i,t},\! \!\bbW_{i,t} ; (\bbX_{i,t}, \bby_{i,t}) )$ and $ \sum_{i=1}^{N} P(\hat{\bby}_{i,t} = \bby_{i,t})/N $ versus iteration $t$, both of which decline faster in smaller networks. Figures  \ref{subfigc_vary_n}-\ref{subfigd_vary_n} show that network disagreement in terms of $RV(\bar{\bbD}_{i,t})$ and $RV(\bar{\bbW}_{i,t})$ becomes more stable and declines faster with smaller $N$. Algorithm performance in moderate sized networks is comparable to the centralized case.}\vspace{-4mm}
\label{fig:vary_n_brodatz} \end{figure*}

To do so, we consider recent advances that make use of sparse representations which have yielded key insights in pattern recognition\cite{Wright:2009:RFR:1495801.1496037}. We select $f$ in \eqref{eq:projection} as a sparse coding problem via the elastic-net \eqref{eq:elastic_net} with the Euclidean distance $\tdf (\bbalpha_t, \bbD; \bbx_t) =\|\bbx_t - \bbD \bbalpha_t \|/2$. Using this loss, \eqref{eq:projection} may be efficiently computed via least angle regression \cite{Efron04leastangle}, whereby we study the performance of Algorithm \ref{alg:d4l} on a multi-class texture classification problem.

We conduct numerical experiments on the Brodatz dataset \cite{Brodatz1966} for a variety of network sizes and topologies. In the case of studying the impact of network size, we also compare the algorithm performance to the centralized case, i.e. $N=1$. Moreover, we consider the case where each agent observes training examples which are incomplete random subsets of the total class labels, yet aims to learn a classifier over all possible classes. We use the numerical insight gained in this section as design principles for the robotic experiments in Section \ref{sec:robots}. 

We restrict ourselves to $C=4$ class labels  $\{${\it grass, bark, straw, herringbone\_weave }$\}$ in the Brodatz texture database, sample images of which are shown in Figure \ref{fig:exampledata}. This data subset consists of one grayscale image per texture, yielding four $512\times512$ images in total consisting of $1,820$ overlapping patches of size $24\times24$. 
\vspace{-5mm}
\subsection{Feature Generation}\label{subset:features}
Inspired by the two-dimensional texton features discussed in \cite{Leung1999}, we generate texture features to classify, $\tbalpha^*$, as the sum of the sparse dictionary representations of sub-patches.
That is, we classify each image patche of size $24$-by-$24$ by first extracting the nine non-overlapping $8$-by-$8$ sub-patches $\bbx^{(i)}$ within it.  We vectorize (column-major order) each sub-patch, subtract off the sample mean, and divide by its norm such that  $\ell_2$ norm so that $\bbx^{(i)}$ is zero-mean and has unit $\ell_2$ norm.  Stacking these sub-patches column wise in a matrix $\mathbf{X} = \left[ \bbx^{(1)}; \cdots ;\bbx^{(9)} \right]$, we compute the aggregate sparse coding $\tbalpha^*_{i,t}$ at agent $i$ at time $t$ according to
%
\begin{equation}\label{eq:subpatch_average}
\tbalpha^*(\bbX_{i,t},\bbD_{i,t}) = \sum_{l=1}^9 {\bbalpha}^*(\bbD_{i,t};\bbx_{i,t}^{(l)}) \; ,
\end{equation}
which implies that the local stochastic gradient of the dictionary  $\nabla_{\bbD_i} \tilde{h}_{i,t} (\bbD_{i,t}, \bbw_{i,t} ; (\bbX_{i,t}, \bby_{i,t}))$ is the sum of contributions  $\nabla_{\bbD_i} {h}_{i,t} (\bbD_{i,t}, \bbw_{i,t} ; (\bbx_{i,t}, \bby_{i,t}))$ from each sub-patch.
%
%
%
\begin{figure*}
\begin{subfigure}{.5\columnwidth}
\includegraphics[width=\linewidth, height = 0.5\linewidth]
                {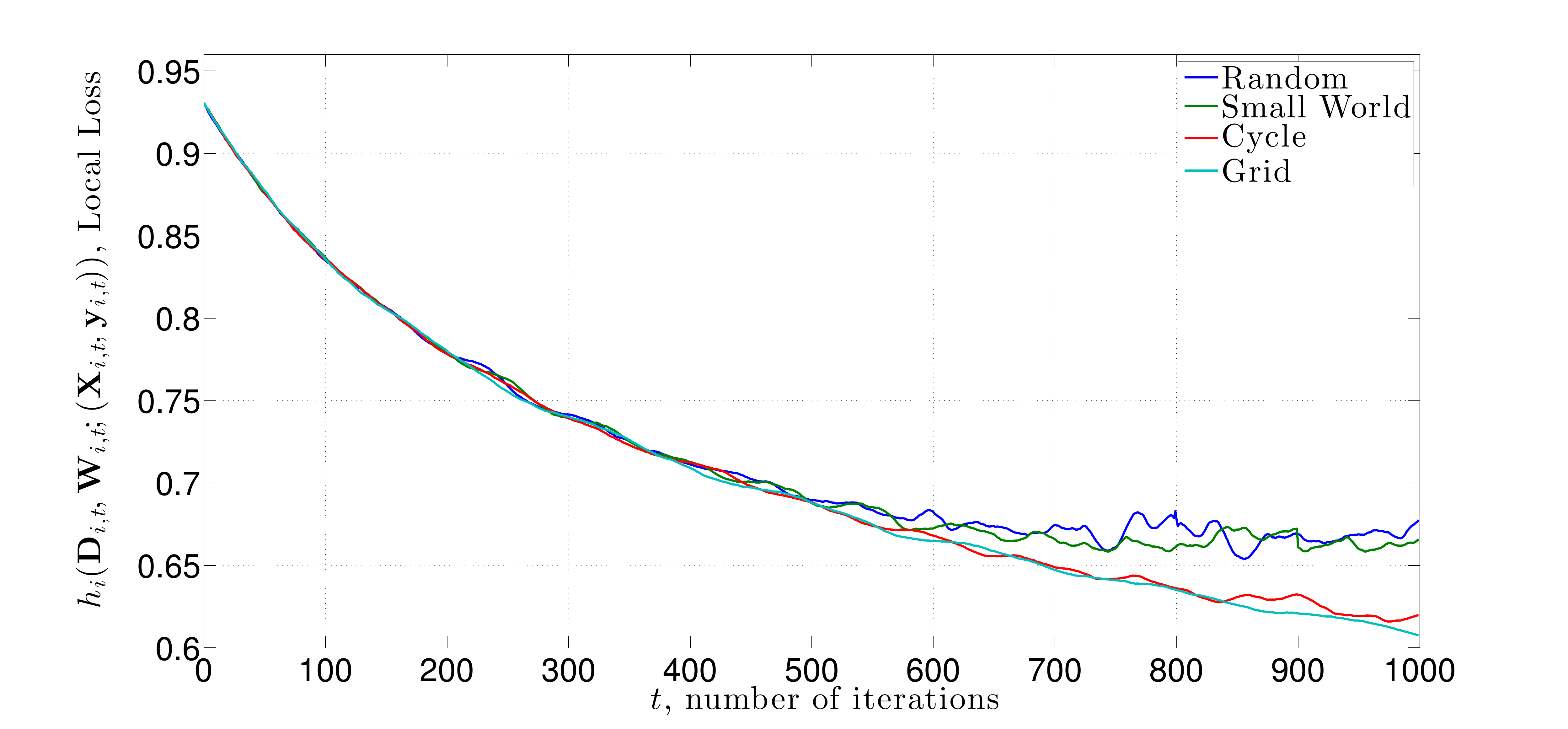}\vspace{-2mm}
\caption{Local loss versus iteration $t$}
\label{subfiga_vary_network}
\end{subfigure}
\begin{subfigure}{.5\columnwidth}
\includegraphics[width=\linewidth,height = 0.5\linewidth]
                {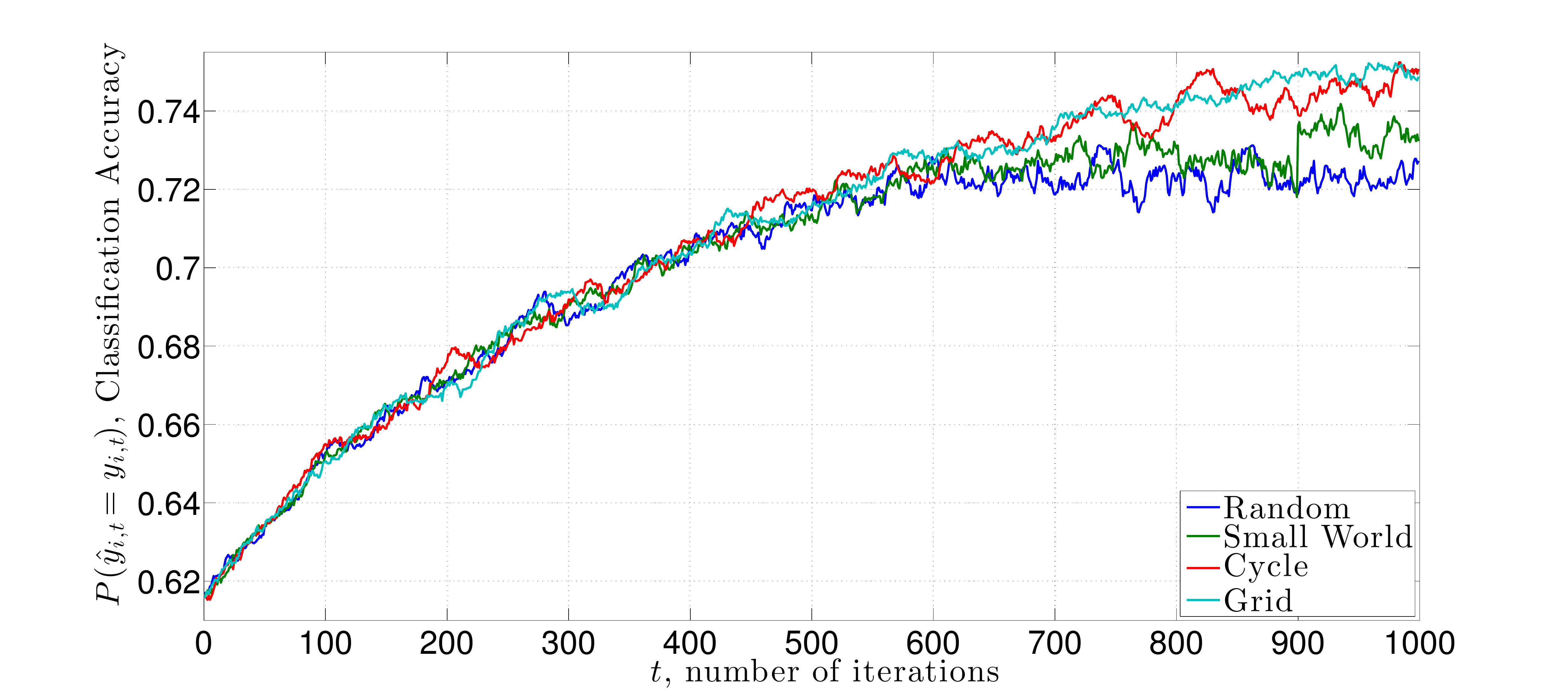}\vspace{-2mm}
\caption{Classification accuracy vs. iteration $t$}
\label{subfigb_vary_network}
\end{subfigure} \\
\begin{subfigure}{.5\columnwidth}
\includegraphics[width=\linewidth,height = 0.5\linewidth]
                {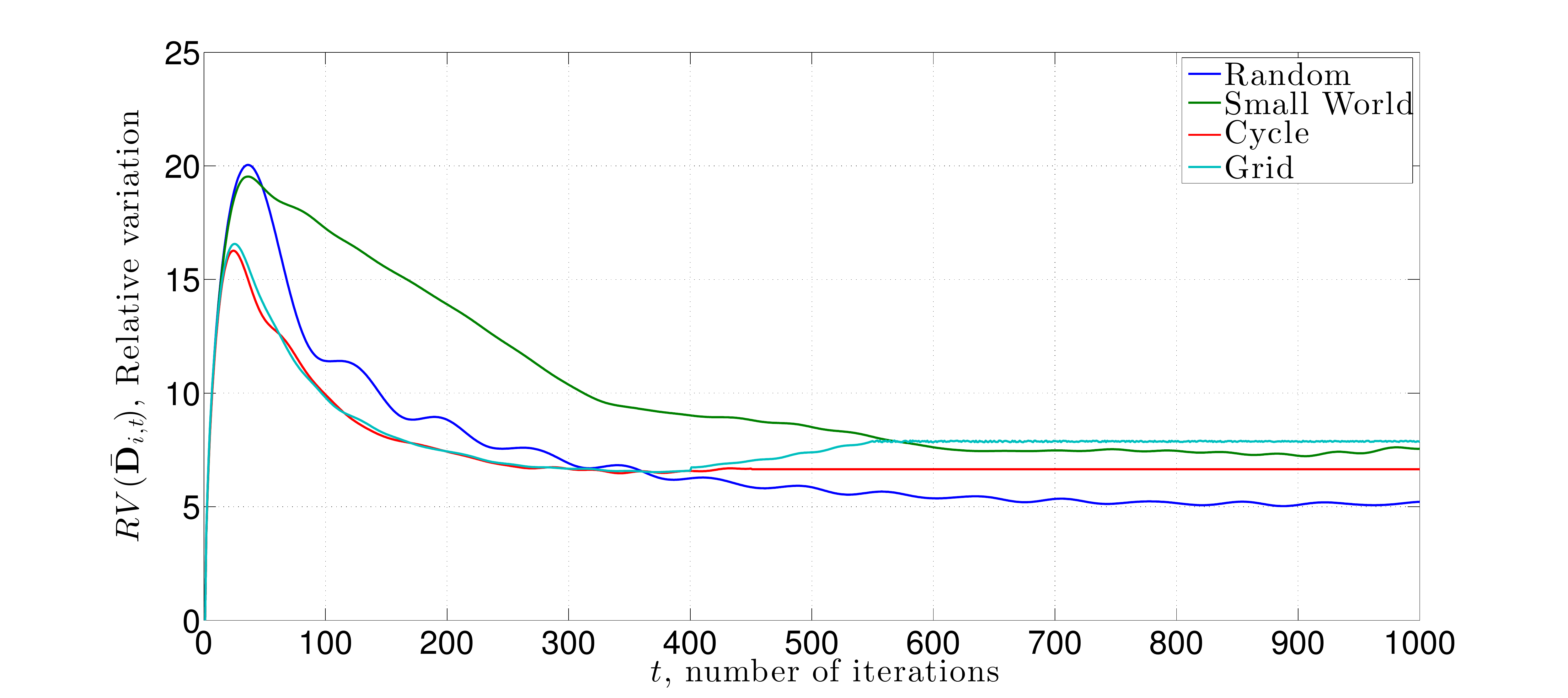}\vspace{-2mm}
\caption{Relative variation $\RV(\bar{\bbD}_{i,t})$ vs. iteration $t$}
\label{subfigc_vary_network}
\end{subfigure}%
\begin{subfigure}{.5\columnwidth}
\includegraphics[width=\linewidth,height = 0.5\linewidth]
                {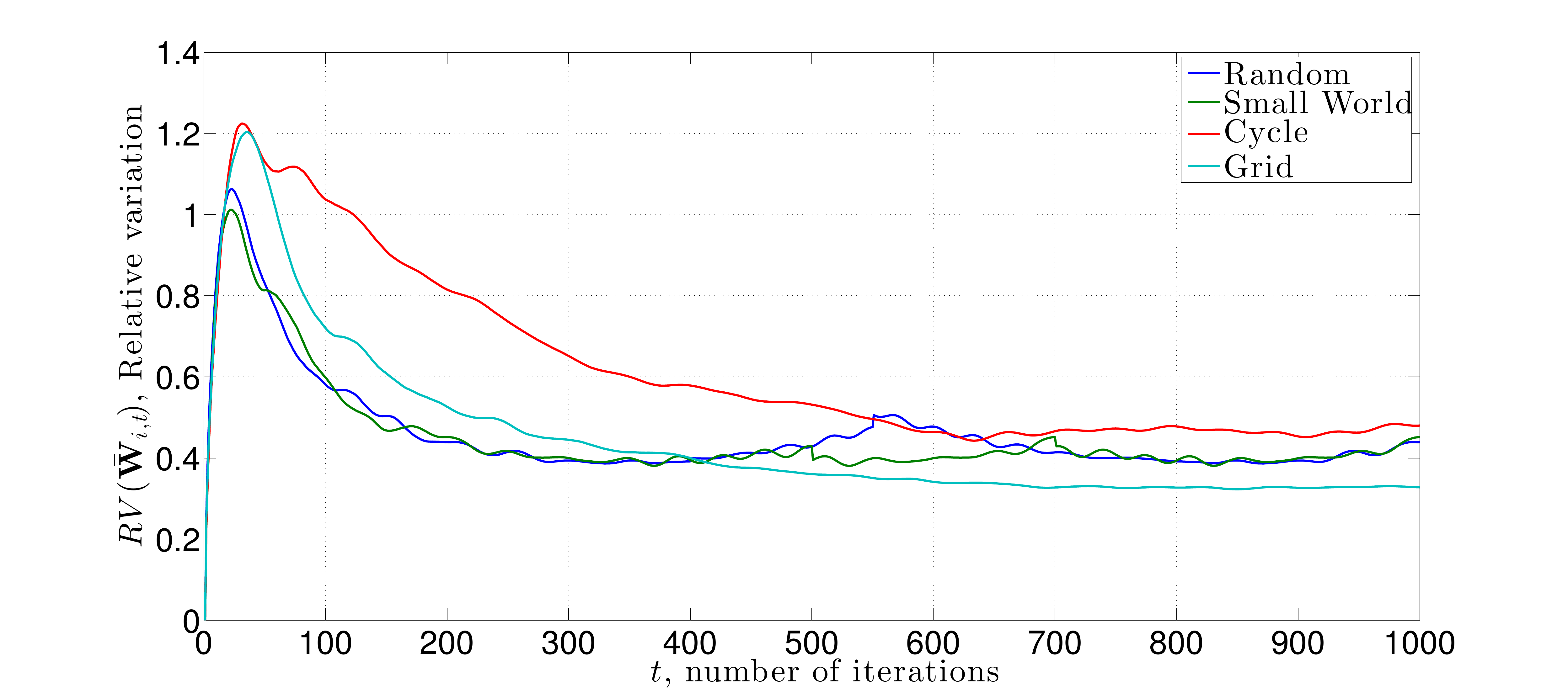}\vspace{-2mm}
\caption{Relative variation $\RV(\bar{\bbW}_{i,t})$ vs. iteration $t$}
\label{subfigd_vary_network}
\end{subfigure}%
\caption{D4L run on $N=20$ node cycle, grid, random and small world networks, where edges are generated randomly between agents with probability $\rho=0.2$ in the later two. Figure \ref{subfiga_vary_network}-\ref{subfigb_vary_network} show $h_{i} (\bbD_{i,t},\! \!\bbW_{i,t} ; (\bbX_{i,t}, \bby_{i,t}) )$ and $ \sum_{i=1}^{N} P(\hat{\bby}_{i,t} = \bby_{i,t})/N $, respectively, over iteration $t$ for an arbitrarily chosen agent $i\in V$. Learning slows and numerical oscillations become more prevalent in networks with random connectivity patterns. Structured deterministic networks such as grids and cycles have larger diameter than small world and random networks,  yet achieve superior performance. Figures \ref{subfigc_vary_network}-\ref{subfigd_vary_network} shows that the agents reach consensus in terms of $\RV(\bar{\bbD}_{i,t})$ and $\RV(\bar{\bbW}_{i,t})$ at comparable rates across the different network topologies.}\vspace{-5mm}
\label{fig:vary_network} \end{figure*}
\subsection{Loss Function and Performance Metrics}\label{subsec:loss}
We cast texture classification as a multi-class logistic regression problem in which agent $i$ receives signals $\bbx_{i}$ and is charged with outputting a decision variable  $\bby_{i,} \in \{0,1\}^C$ where $C$ is the number of classes. Each component $y_{i,c}$ of the vector $\bby_{i}\in \{0,1\}^C$ is a binary indicator of whether the signal falls into class $c$. 
The local loss $h_{i}$ is the negative log-likelihood of the probabilistic model\cite{Murphy2012}, stated as
\begin{align}\label{eq:multi_logistic}
h_{i} (\bbD_i,\! \!\bbW_i ; (\bbX_i, \bby_i) )
	 &= \log\left( \sum_{c=1}^C e^{\bbw_{i,c}^T\tbalpha^*_i + w^0_{i,c}} \! \!\right)  \\
	 &\qquad- \!\sum_{c=1}^C\! \Big( y_{i,c} \bbw_{i,c}^T \tbalpha_i^* \! +\!w^0_{i,c} \Big)\!+\! \xi \| \bbW_i \|^2_F\; , \nonumber
\end{align}
where the activation functions $g_c(\tbalpha_i^*) = e^{\mathbf{w}_{i,c}^T\tbalpha_i^* + w^0_{i,c}}$, are computed using the $c^\text{th}$ column $\bbw_c$ of the weight matrix $\bbW_i \in \mathbb{R}^{(k+1) \times C}$.  Moreover, $w^0_{i,c}$ is a bias term for each class $c$. Recall that $h_{i}$ is implicitly a function of $\bbD_i$ since the aggregate sparse coding $\tbalpha_i^*$ depends on the dictionary over which it is represented. Moreover, we add $l_2$ regularization to prevent over-fitting.

To ensure identifiability, every element of the last column of $\bbW_i$ is set to zero. With $\mathbf{W}_i$, the probability that $\tbalpha^*_i$ belongs to class $c$ is given by $g_c(\tbalpha^*_i)/\sum_{c'}g_{c'}(\tbalpha^*_i)$, the classification decision is made via maximum-likelihood estimation, i.e. $\tilde{c}=\argmax_c g_c(\tbalpha^*_i)/\sum_{c'}g_{c'}(\tbalpha^*_i)$ is the only nonzero entry of $\bby_{i,t}$.

Besides the local loss $h_{i}$ which we known converges to a KKT point as a consequence of Theorem \ref{thm1}, we also study the network average classification accuracy $ \sum_{i=1}^{N} P(\hat{\bby}_{i,t} = \bby_{i,t})/N $ at each iteration. Here $\bby_{i,t}$ denotes the true texture label, $\hby_{i,t}$ denotes the predicted label, and $ P(\hat{\bby}_{i,t} = \bby_{i,t})$ represents the empirical classification rate on a fixed test set of size $\tilde{T}=4096$. We also consider the relative variation of the average classifiers, stated as 
\begin{equation}\label{eq:disagreement}
\RV(\bar{\bbW}_{i,t} )= \frac{1}{N}\sum_{j=1}^N \| \bar{\bbW}_{i,t} - \bar{\bbW}_{j,t} \|_F,
\end{equation}
 where $\bar{\bbW}_{i,t}= \sum_{s=1}^t \bbW_{i,s}/t $ which quantifies how far individual agents' classifiers are from consensus. We consider time averages $\bar{\bbW}_{i,t}$ instead of the plain estimates $\bbW_{i,t}$ because the latter tend to oscillate around the stationary point $\bbW^*$ and agreement between estimates of different agents is difficult to visualize.
 \vspace{-3mm}
\subsection{Implementation Details}\label{subset:design}

%
%

{\it (i) Dictionary Size}
 To select a dictionary of the appropriate size, we investigate its effect on classification accuracy on the Brodatz textures. Because the number of atoms $k$ in the dictionary will similarly affect both the centralized and decentralized algorithms, we conduct out this experiment for the centralized ($N=1$) algorithm due to computational expedience.
As in \cite{10.1109/TPAMI.2011.156}, we find that increasing the size of the dictionary led to better classifier performance. However, because of diminishing performance improvement with increasing $k$ beyond a certain level, we select $k=128$ in all subsequent experiments. In theory, one could make $k$ arbitrarily large, but as $k$ increases so do the computational demands of the algorithm. We show the initialized and final $128$-element, $8$-by-$8$ patch dictionaries in Figure \ref{fig:initdictionary}.

{\it (ii) Mini-Batching}
In our implementation of D4L, we adopted a mini-batching procedure: at each iteration, we replace the single labeled patch with a small batch  of $\hat{T}=200$ randomly-drawn labeled patches.
The procedure for generating this batch is as follows: for each patch, a label is first drawn uniformly at random from the set of all possible labels.
Then, the patch is selected uniformly at random from the set of all patches with that label.
We then compute the dictionary and classifier gradient values for the iteration by averaging the gradient values generated by each individual patch within the mini-batch.
Practically, this process reduces the variance of the local stochastic gradients, which often empirically yields improved convergence. 

{\it (iii) Initialization}
We initialized $\mathbf{D}$ using unsupervised dictionary learning for a small set of randomly-drawn initialization data \cite{Mairal:2010:OLM:1756006.1756008}.
We then used the labels and the dictionary representations of the data to initialize the classifier parameters $\mathbf{W}$. All experiments are run from this common initialization.

{\it (iv) Regularization and Step-size Selection}
The D4L algorithm requires several parameters to be specified.
Following \cite{10.1109/TPAMI.2011.156}, we used $\zeta_1=0.125$, $\zeta_2=0$, $\xi=10^{-9}$. We also adopted the learning-rate selection strategy discussed in \cite{10.1109/TPAMI.2011.156}, which is to select the initial step-size $\epsilon$ by implementing a grid search over a fixed small number of iterations ($\tilde{T}=200$) and selecting the one that minimized cross-validation error. We set $\epsilon_t = \min(\epsilon,\epsilon t_0/t)$, where  $t_0 = T/2$. This selection amounts to using a step-size of $\epsilon$ for the first half of the iterations after which we follow a $1/t$ annealing rate for those that remain, enforcing convergence. Moreover, we select the gradient projection threshold as $\eta=.25$.
%
\begin{figure*}
\begin{subfigure}{.5\columnwidth}
\includegraphics[width=\linewidth, height = 0.5\linewidth]
                {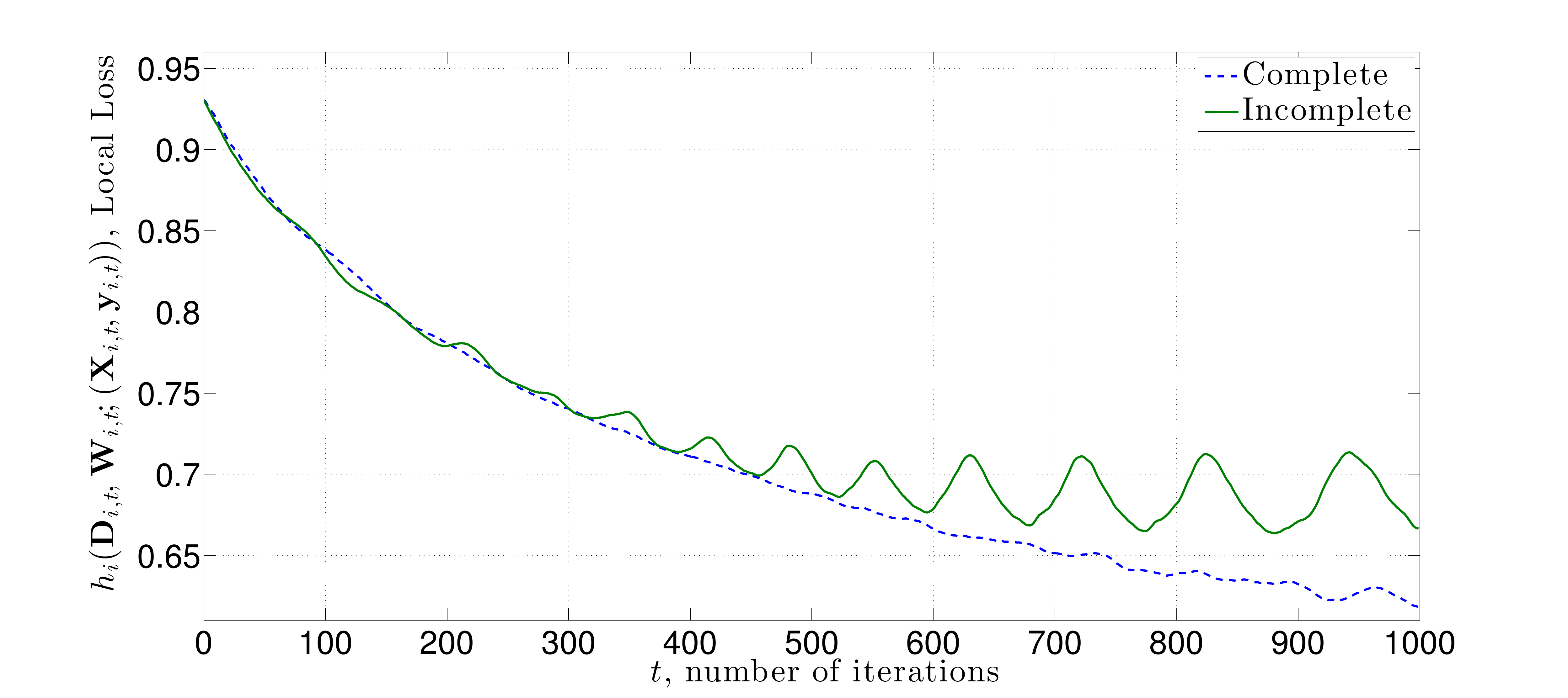}
\caption{Local loss versus iteration $t$}
\label{subfiga_incomplete}
\end{subfigure}
\begin{subfigure}{.5\columnwidth}
\includegraphics[width=\linewidth,height = 0.5\linewidth]
                {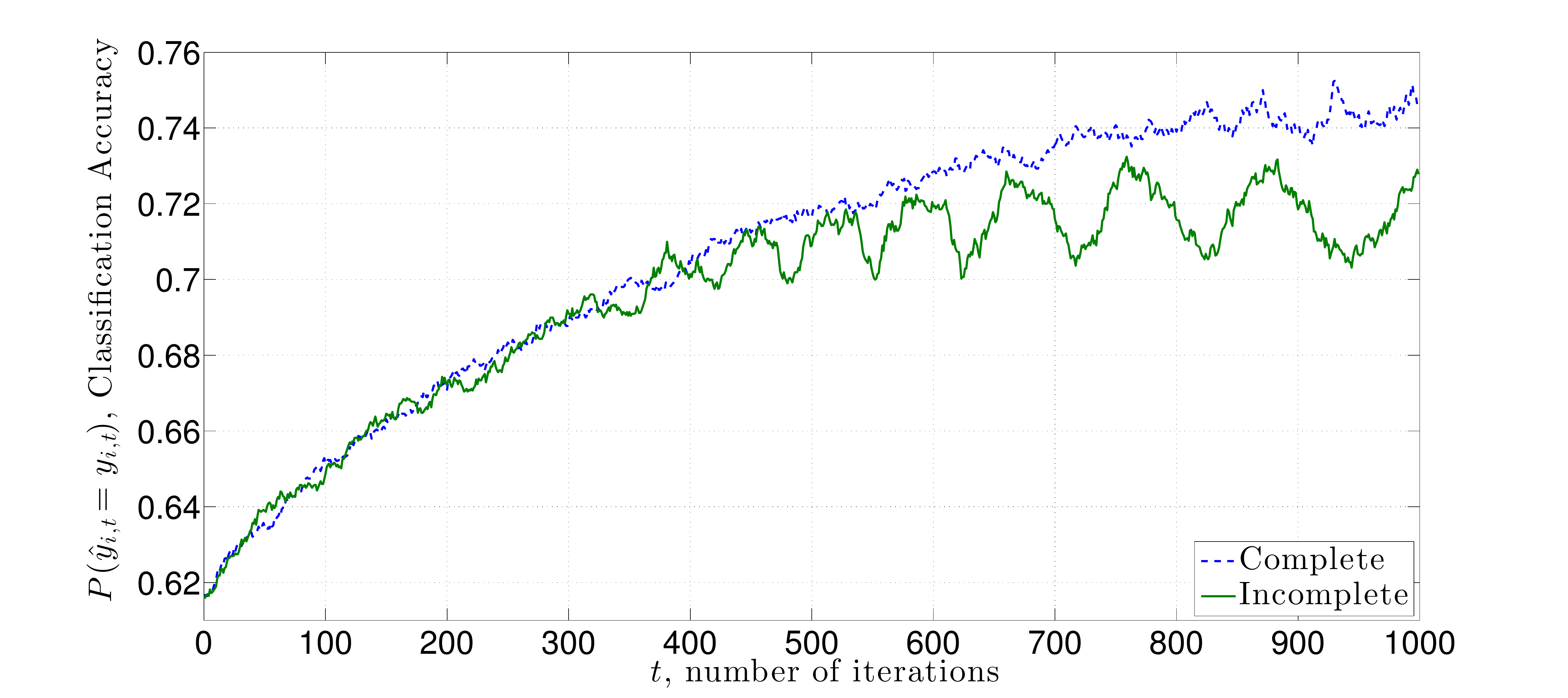}
\caption{Classification accuracy vs. iteration $t$}
\label{subfigb_incomplete}
\end{subfigure}
\begin{subfigure}{.5\columnwidth}
\includegraphics[width=\linewidth,height = 0.5\linewidth]
                {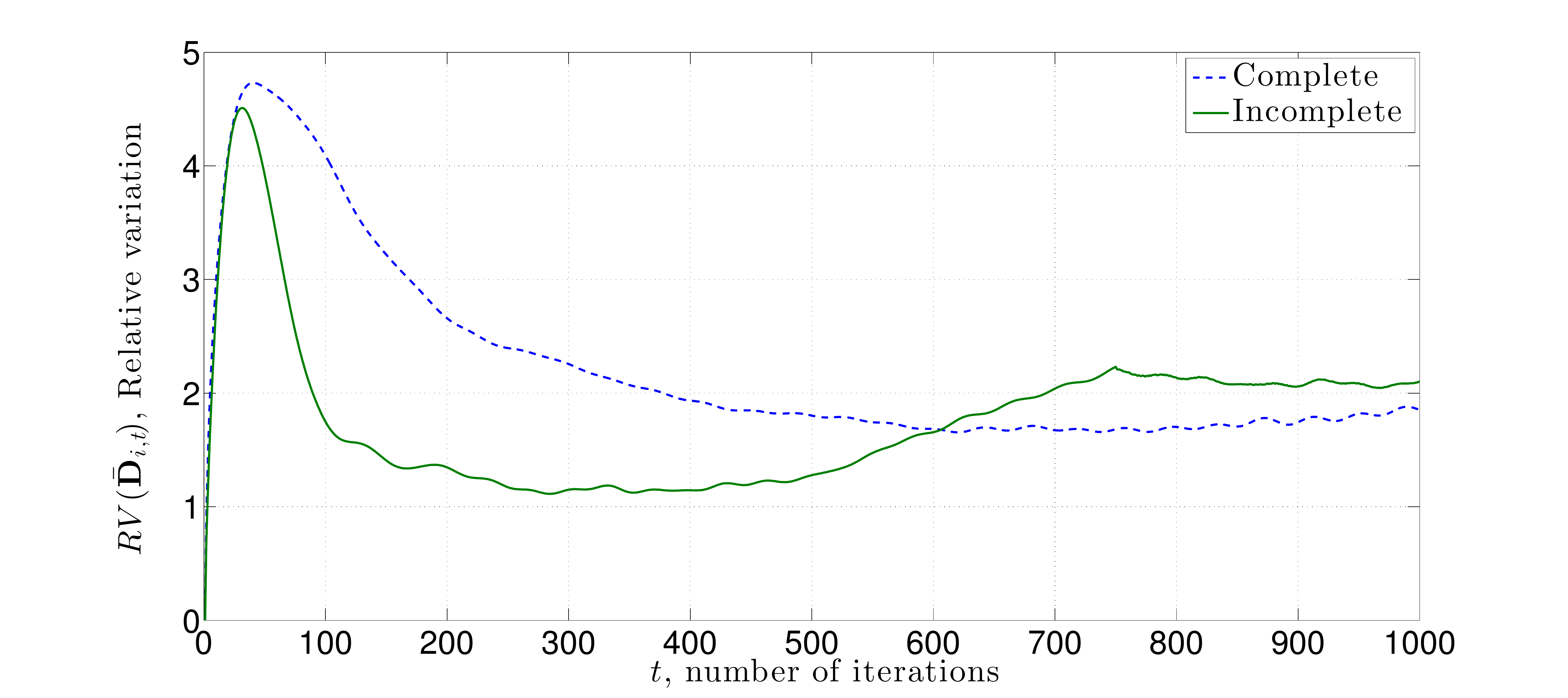}
\caption{Relative variation $\RV(\bar{\bbD}_{i,t})$ vs. iteration $t$}
\label{subfigc_incomplete}
\end{subfigure}%
\begin{subfigure}{.5\columnwidth}
\includegraphics[width=\linewidth,height = 0.5\linewidth]
                {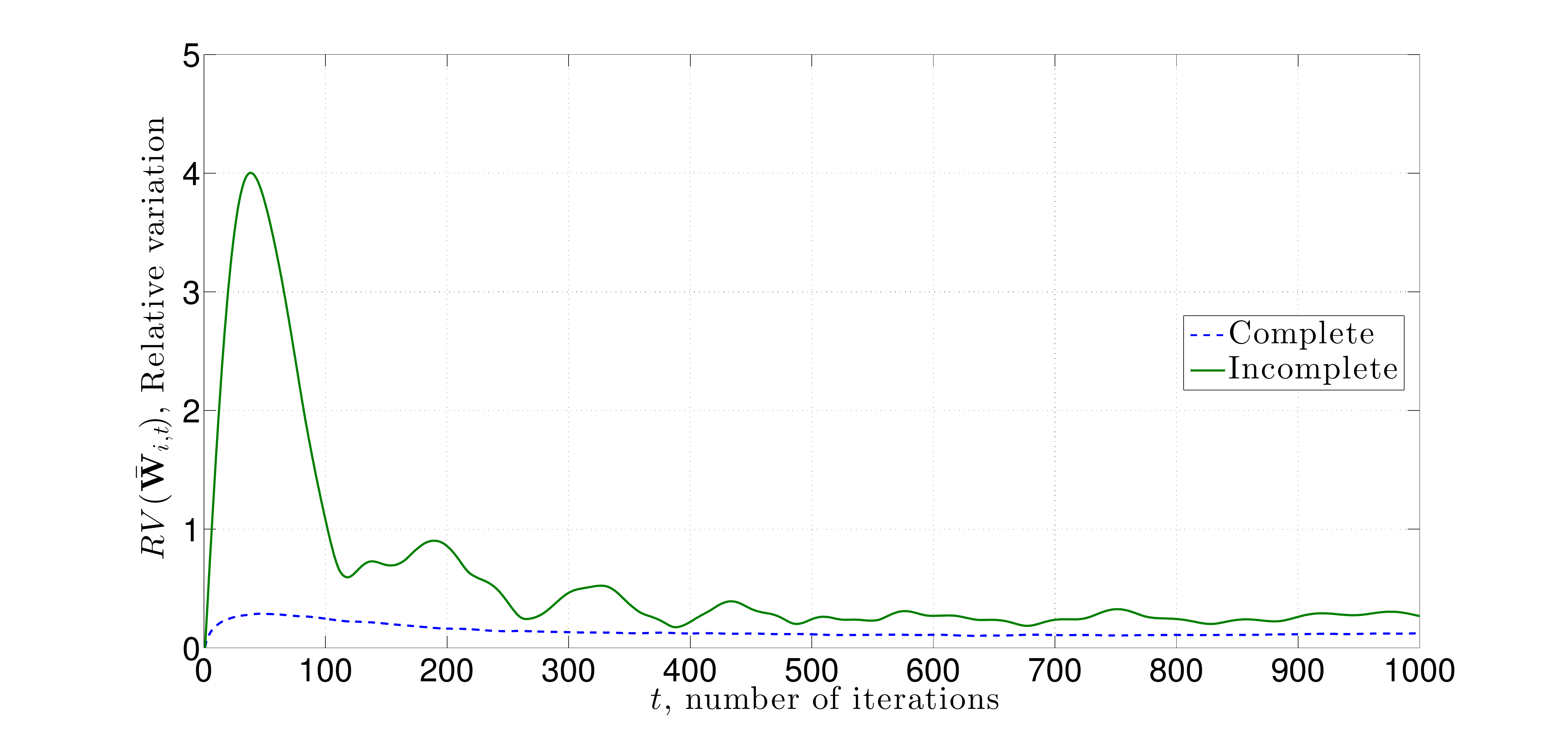}
\caption{Relative variation $\RV(\bar{\bbW}_{i,t})$ vs. iteration $t$}
\label{subfigd_incomplete}
\end{subfigure}%
\caption{Algorithm \ref{alg:d4l} run on a $N=10$ node random networks, where edges are generated randomly between agents with probability $\rho=0.2$. "Incomplete" refers to the case where each agent observes training examples that comprise a random incomplete subset of the total data labels. Figure \ref{subfiga_vary_network}-\ref{subfigb_vary_network} show $h_{i} (\bbD_{i,t},\! \!\bbW_{i,t} ; (\bbX_{i,t}, \bby_{i,t}) )$ and $ \sum_{i=1}^{N} P(\hat{\bby}_{i,t} = \bby_{i,t})/N $, respectively, over iteration $t$ for an arbitrarily chosen agent $j\in V$, as compared with the case where this agent observes training examples from all classes. Observe that learning is still achieved for this more challenging context, yet the algorithm exhibits increased oscillatory behavior and decreased accuracy for the incomplete case. Figures \ref{subfigc_vary_network}-\ref{subfigd_vary_network} shows that the agents reach consensus in terms of $\RV(\bar{\bbD}_{i,t})$ and $\RV(\bar{\bbW}_{i,t})$ at comparable rates for complete cases and incomplete cases, albeit with more oscillations in the later. Moreover, the algorithm still converges despite the instability in $\RV(\bar{\bbD}_{i,t})$. }\vspace{-4mm}
\label{fig:incomplete} \end{figure*}
We note here that, due to the non-convexity of the objective, the algorithm may diverge if $\epsilon$ is too large. This follows from the fact that convergence guarantees for stochastic gradient algorithms in non-convex settings only occur in cases where a diminishing step-size mitigates the stochastic approximation error, which may not occur if the step-size is too large.
Moreover, we have experimentally observed that values of $\epsilon$ which avoid this behavior are smaller than effective values for the centralized version by an order of magnitude or more. Consequently,  when comparing D4L to its centralized counterpart, we select $\epsilon$ that yield convergence for both settings, i.e., the smaller values appropriate for D4L. For the Brodatz dataset, we found that $\eps=0.05$ led to convergence in all cases. In all subsequent results in Section \ref{subset:results}, we run the algorithm for a total of $\bar{T}=20$ sample paths, and report the results in terms of the empirical mean realization.
%
\subsection{Results on Texture Database}\label{subset:results}

{\it (i) Network Size}
To investigate the dependence of the convergence rate in Theorem \ref{thm1} on the network size $N$ we run Algorithm \ref{alg:d4l} for problem instances with $N=1$ (centralized), $N=10$, and $N=100$ nodes. For the later two cases, connections between nodes are random, with the probability of two nodes being connected set to $\rho=0.2$. Because such randomly generated networks are not guaranteed to be connected, we repeatedly generate networks according to this rule and take the first which is connected. We repeat this process until we obtain a network which has a fixed connectivity ratio $\varrho=0.2$, which implies that the average degree for individual nodes as the network size grows. In this experiment, each agent observes training examples from all label classes. We note that the centralized case $N=1$ is comparable to existing state of the art methods.

Figure \ref{fig:vary_n_brodatz} shows the results of this numerical experiment for a randomly selected agent in the network. In Figure \ref{subfiga_vary_n}, we show $h_{i} (\bbD_{i,t},\! \!\bbW_{i,t} ; (\bbX_{i,t}, \bby_{i,t}) ) $ over iteration $t$. Observe that as $N$ increases, the log-likelihood $h_{i} (\bbD_{i,t},\! \!\bbW_{i,t} ; (\bbX_{i,t}, \bby_{i,t}) )$ declines at comparable rates for networks of moderate size, yet it significantly slower for the $N=100$ node network. To be specific, both the centralized and $N=10$ node network achieve $h_{i} (\bbD_{i,t},\! \!\bbW_{i,t} ; (\bbX_{i,t}, \bby_{i,t}) )\leq 0.62$ by $T=1000$, while the $N=100$ node network remains at $0.77$ over its run. We may observe this performance discrepancy more concretely in Figure \ref{subfigb_vary_n} which shows the classification accuracy on a fixed test set over iteration $t$. Observe that the centralized algorithm achieves an accuracy near $76\%$, whereas the decentralized methods achieve an accuracy of $75\%$ and $67\%$ for the $N=10$ and $N=100$ node networks by $T=1000$ iterations, respectively.

In Figure \ref{subfigc_vary_n} we investigate how far the agents are from consensus as measured by  $\RV(\bar{\bbD}_{i,t})$ over iteration $t$. Observe that the network experiences greater disagreement, as measured with $\RV(\bar{\bbD}_{i,t})$, with larger $N$. Trivially $\RV(\bar{\bbD}_{i,t})=0$ for the centralized case, but for the $N=10$ and $N=100$ node networks the algorithm achieves $\RV(\bar{\bbD}_{i,t})\leq1.9$ by $t\geq 417$ and  $\RV(\bar{\bbD}_{i,t})\leq 35$ by $T= 1000$ respectively. Thus in larger networks information diffuses more slowly. Moreover, the agreement constraints are more difficult to satisfy and delay the convergence to stationarity. This difference in consensus achievement may also be observed in Figure \ref{subfigd_vary_n}, which shows $\RV(\bar{\bbW}_{i,t})$ over iteration $t$. That is, for the $N=10$ and $N=100$ node networks the algorithm achieves $\RV(\bar{\bbW}_{i,t})\leq1.3\times 10^{-1}$ by $t\geq 312$ and  $\RV(\bar{\bbW}_{i,t})\leq 3$ for $t\geq 400$ respectively. Thus we observe an order of magnitude difference in the relative variation between the $N=10$ and $N=100$ node networks for both the dictionary and model parameters.
%
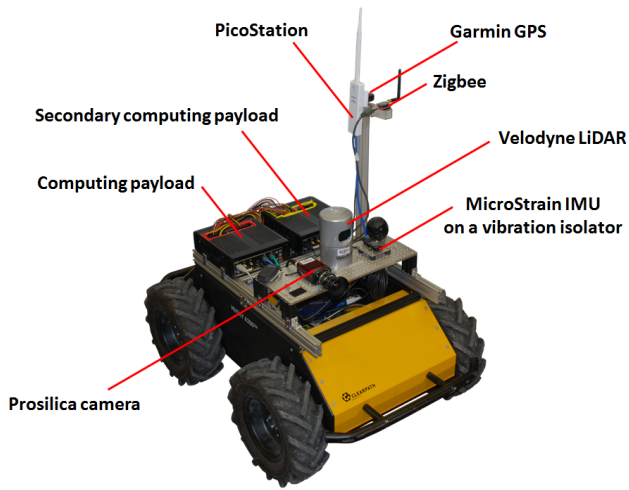
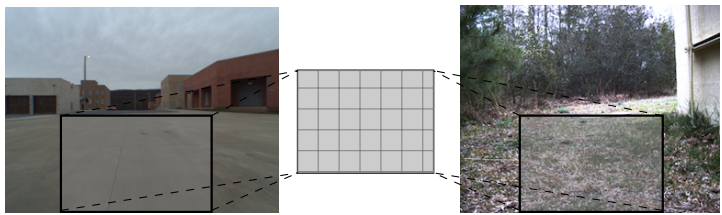
\begin{figure*}
\centering
\begin{subfigure}{.45\columnwidth}
	\includegraphics[width=\linewidth]{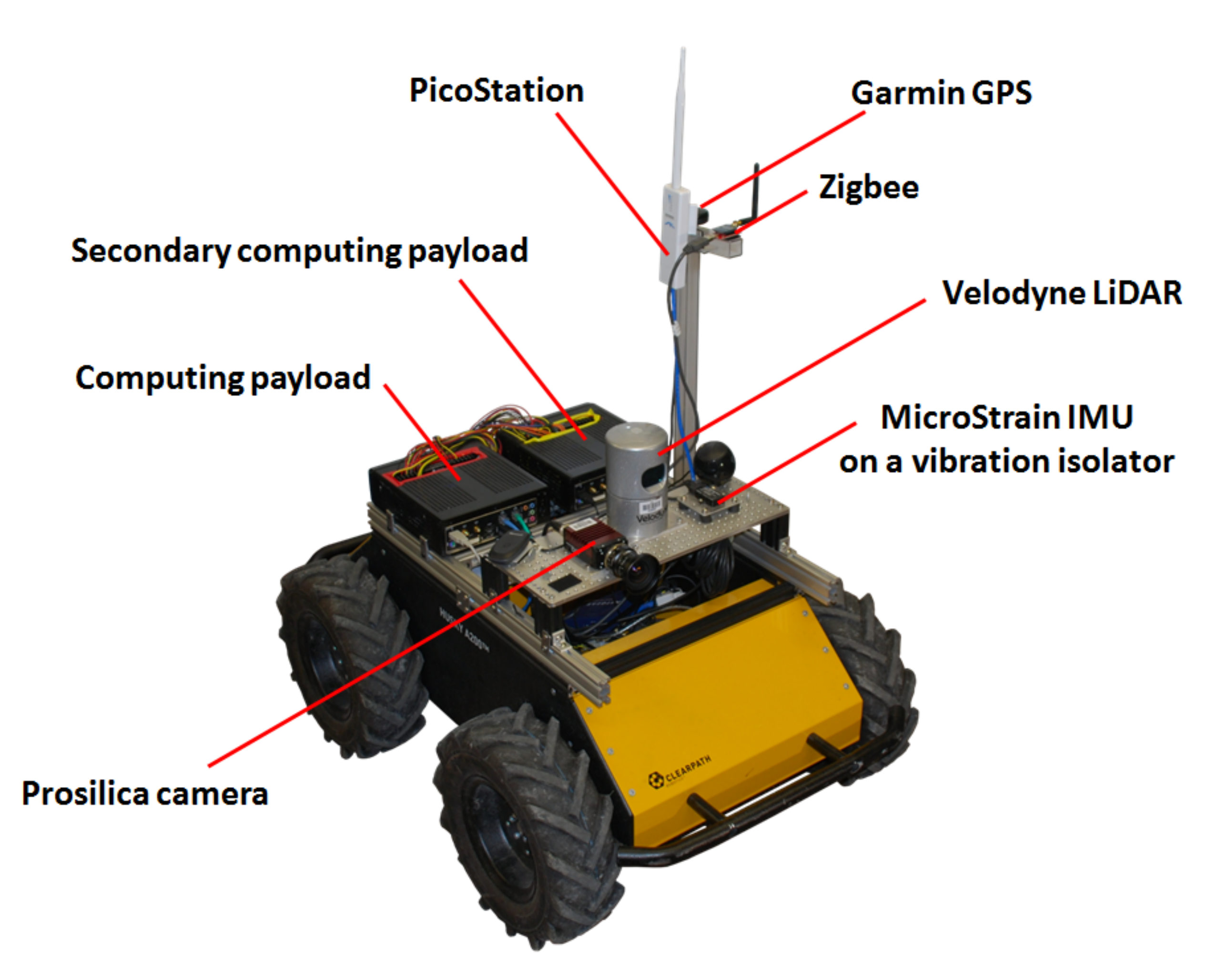}\vspace{-2mm}
		\caption{Clearpath Husky robot.} 
		\label{fig:platform}
\end{subfigure} 
\hspace{5mm}
\begin{subfigure}{.5\columnwidth}
\begin{tikzpicture}
    \node[anchor=south west,inner sep=0] at (0,0) {\includegraphics[width=0.38\linewidth]{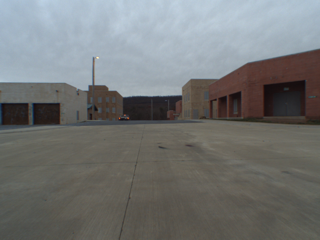}};
    \draw[black, thick] (.75,.02) rectangle (2.7525,1.3);
        \draw[black, thick, opacity=0.5, fill=gray] (.75,.02) rectangle (2.7525,1.3);
    \draw[black, thick] (3.9,.54) rectangle (5.7,1.9);
    \draw[step=.278cm,gray,very thin,fill=black!20!white] (3.9,.54) grid (5.7,1.9)  rectangle (3.9,.54);
    \draw[step=.4cm,black,dashed] (2.7525,1.3) -- (3.9,1.9);
\draw[step=.4cm,black,dashed] (.75,1.3) -- (3.9,1.9);
\draw[step=.4cm,black,dashed] (.75,.04) -- (3.9,.54);
\draw[step=.4cm,black,dashed] (2.7525,.04) -- (3.9,.54);
        \node[anchor=south west,inner sep=0] at (6.05,0) {\includegraphics[width=0.36\linewidth]{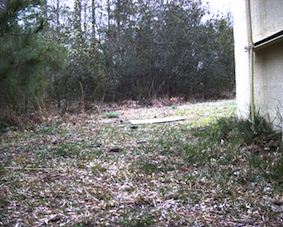}};
            \draw[black, thick] (6.85,.02) rectangle (8.75,1.3);
            \draw[black, thick,opacity=0.5, fill=gray] (6.85,.02) rectangle (8.75,1.3);
                \draw[step=.4cm,black,dashed] (6.85,1.3) -- (5.7,1.9);
\draw[step=.4cm,black,dashed] (8.75,1.3) -- (5.7,1.9);
\draw[step=.4cm,black,dashed] (6.85,.04) -- (5.7,.54);
\draw[step=.4cm,black,dashed] (8.75,.04) -- (5.7,.54);
\end{tikzpicture} 
\caption{Image extraction for learning a dictionary tailored to skid prediction.}\label{fig:examplerobot}
\end{subfigure}%

\caption{The Husky robot in Figure \ref{fig:platform} measures state information via an on-board IMU and communicates with other robots using a Zigbee radio.
Sample images from the IRA dataset associated with an $N=3$ node network of Husky robots moving around a cluttered urban setting are shown in Figure \ref{fig:examplerobot}. The platforms extract the sub-patch of the image associated with the ground to which they will travel at the next time slot.}\vspace{-5mm} 
\label{fig:test} \end{figure*}

{\it (ii) Network Topology and Diameter}
We study the dependence of the convergence rate of Algorithm \ref{alg:d4l} in Theorem \ref{thm1} on the network topology by fixing the network size to $N=20$ and running \eqref{eq:sp_dict} - \eqref{eq:sp_dual_w} over random graphs, small world graphs, cycles, and grids. In the first two, the probability that node pairs are randomly connected is fixed at $\rho=0.2$. Again, we repeatedly generate these random networks and select the first realization which is connected for our simulation. The latter two are deterministically generated. A cycle is a closed directed chain of nodes. Grids are formed by taking the two-dimensional integer lattice of size $\sqrt{N}\times\sqrt{N}$, with $\sqrt{N}$ rounded to the nearest integer. Connections are drawn between adjacent nodes in the lattice as well as between remainder nodes at the boundary. Cycles, grids and random networks have progressively larger number of connections per node and smaller diameter. Random networks have small degree and small diameter; see \cite{watts,west_introduction_2000}.

We present the results of this experiment in Figure \ref{fig:vary_network}. In Figure \ref{subfiga_vary_network}, we plot  $h_{i} (\bbD_{i,t},\! \!\bbW_{i,t} ; (\bbX_{i,t}, \bby_{i,t}) ) $ over iteration $t$. Observe that the rate at which $h_{i} (\bbD_{i,t},\! \!\bbW_{i,t} ; (\bbX_{i,t}, \bby_{i,t}) )$ decreases is comparable across the different networks, yet we can differentiate the learning achieved in the different settings by the benchmark $h_{i} (\bbD_{i,t},\! \!\bbW_{i,t} ; (\bbX_{i,t}, \bby_{i,t}) ) \leq 6.5\times10^{-1}$. To surpass this bound, the algorithm requires $t=700$ iterations for grids and cycles, whereas the random and small world networks oscillate around the interval $[0.65,0.70]$ for $t\geq 650$. This indicates that structured deterministic networks are an easier setting for finding good signal representations in a decentralized manner. This insight is corroborated in the plot of classification performance over iteration $t$ in Figure \ref{subfigb_vary_network}. Observe that the algorithm achieves an accuracy near $75\%$ for cycle and grid networks as compared to $72\%$  and $73\%$ for random and small world networks, respectively. 

We study how the network topology impacts the algorithm's convergence to consensus in  \ref{subfigc_vary_network}, where we plot  $\RV(\bar{\bbD}_{i,t})$ over time $t$. Observe that the initial burn-in period is comparable across the different networks except for the cycle, i.e. $\RV(\bar{\bbD}_{i,t})\leq 7$ for $t\geq 300$  the grid, small world, and random networks as compared to $\RV(\bar{\bbD}_{i,t})\leq 10$ for the cycle setting. However, by $t\geq600$, the random and small world networks experience greater numerical oscillations. This difference in convergence to consensus as measured by the relative variation is corroborated in \ref{subfigd_vary_network}, where we plot  $\RV(\bar{\bbW}_{i,t})$ versus iteration $t$. Observe that the cycle yields the slowest convergence rate, yet is more stable than the small world and random networks. Surprisingly, the grid network has superior convergence to consensus both in terms of dictionary and model parameters.

{\it (iii) Complete vs. Incomplete Sampling}
We now turn to studying the performance of D4L in the more challenging context of what we term {\it incomplete sampling}, which refers to the case that each agent in the network observes only training examples from a fixed random subset of the total number of class labels, yet is charged with the task of learning a classifier which can identify all classes. This problem setting is closer to our robotic application of interest, which is discussed in further detail in Section \ref{sec:robots}. We run the algorithm on a $N=10$ node random network with connection probability $\rho=0.2$. We display these results in Figure \ref{fig:incomplete}, where we also plot for comparison results generated in the complete sampling setting.

Figure \ref{fig:incomplete} shows the results of this numerical experiment for a randomly selected agent in the network. In Figure \ref{subfiga_incomplete}, we plot $h_{i} (\bbD_{i,t},\! \!\bbW_{i,t} ; (\bbX_{i,t}, \bby_{i,t}) )$ over iteration $t$. Note that during an initial burn-in period of $t\leq300$ the local losses decline at comparable rates, after which the algorithm experiences greater numerical oscillations and its convergence rate slows for the incomplete case. In particular, for the benchmark $h_{i} (\bbD_{i,t},\! \!\bbW_{i,t} ; (\bbX_{i,t}, \bby_{i,t}) ) \leq 6.7\times 10^{-1}$, the incomplete sampling case requires $t=676$ iterations as compared with $t=592$ for the complete sampling case. This oscillatory behavior and reduced convergence rate may also be observed in the plot of classification performance versus iteration $t$ in Figure \ref{subfigb_incomplete}. Observe that by $t=350$ both cases achieve an accuracy of $70\%$; however, the incomplete sampling oscillates around this benchmark whereas the complete case continues to improve. Succinctly, increased dithering around stationarity occurs when agents observe training examples drawn from incomplete subsets of the total labels.

We study how the incomplete sampling, or the implicit partition of the feature space across the network, impacts the network disagreement in Figure \ref{subfigc_incomplete}, where we  plot $\RV(\bar{\bbD}_{i,t})$ over time $t$. Observe during an initial burn-in period of $t\leq 100$ that the relative variation is smaller in the case of incomplete sampling for the dictionary, yet by $t\geq600$ the complete sampling case more closes enforces consensus. Moreover, $\RV(\bar{\bbD}_{i,t})$ slowly climbs despite the convergence of the algorithm to a neighborhood of a stationary point. We observe a improved constraint slack convergence in the plot of $\RV(\bar{\bbW}_{i,t})$ over time $t$ in Figure \ref{subfigd_incomplete}, i.e. for $t\geq$, $\RV(\bar{\bbW}_{i,t}) \leq 5\times10^{-1}$ for the incomplete sampling case. This suggests that the empirical effect of non-convexity is predominantly limited to the dictionary learning procedure.

%

\vspace{-2mm}
\section{Robotic Experiments}\label{sec:robots}

%
In this section, we make use of the tools developed in Section \ref{sec:alg} such that a team of robots may predict unknown characteristics of the environmental domain in which they are operating in real-time, and the unexpected maneuvers of one robot may be used to prevent similar steering mistakes in another. To do so, we consider a $N=3$ network of Clearpath Husky robots and driving sample trajectories on both grass and pavement, which we control in an open-loop manner with a joystick. We collected images that were sequentially observed by the robotic network at Camp Lejeune, a cluttered urban setting. The platform then extracts the subset of its observed image associated with the patch of ground defining its trajectory at the next time slot. See Figure \ref{fig:examplerobot} for an example of the extracted images taken by a prototypical Husky platform, which is shown in Figure \ref{fig:platform}. Experimentally, we consider cases where each robot observes training examples from both classes (grass and concrete), i.e. all robots drive on both terrains.

We propose using state information for the purpose of supervision. To do so, during the traversal of these trajectories, we record the commanded linear and angular velocities on each robot.  The Husky is equipped with an inertial measurement unit which we use to acquire measurements of the \emph{actual} angular velocity experienced by the platform. The hardware specifications are detailed in Figure \ref{fig:platform}. Image labels received at robot $i$ at time $t$ are denoted by $y_{i,t}\in \{-1,1\}$ are generated by thresholding the difference between the commanded angular velocity $\tbomega_{i,t}$ and the one measured by the IMU $\bbomega_{i,t}$,
%
\begin{equation}\label{eq:label_imu}
y_{i,t} = 2\cdot \bbone\{|\tbomega_{i,t} - \bbomega_{i,t}|\geq \gamma \} -1  \; ,
\end{equation}
where $\bbone$ denotes the indicator function, and $\gamma$ is a threshold on the magnitude of unexpected rotational maneuvers of the platform, which is experimentally set to $\gamma=.5$. For an experimental setup in which the robotic network only operates on concrete and grass, predicting the image label $y_{i,t}$ amounts to predicting which terrain the robots are more likely to skid via a maximum likelihood estimation of the skidding probability. We view this task as an online collaborative logistic regression problem, whereby local losses take the form \eqref{eq:logistic_loss}. See \cite{cKoppelEtal16a} for a related approach. 
We call this data the Integrated Research Assessment for the U.S. Army's Robotics Collaborative Technology Alliance, abbreviated as IRA. The IRA dataset consists of 16 images converted to grayscale of size $320$ by $240$, which consist of $610,528$ overlapping $24$-by-$24$ patches. Moreover, via on-board radios, each robot is able to communicate with all others.\footnote{\noindent We would like to thank Arne Suppe and Luis Navarro-Serment from Carnegie Mellon University for the use of the IRA data which allowed us to execute a field implementation.}

\begin{figure}
	\centering
	\includegraphics[width=0.45\textwidth]{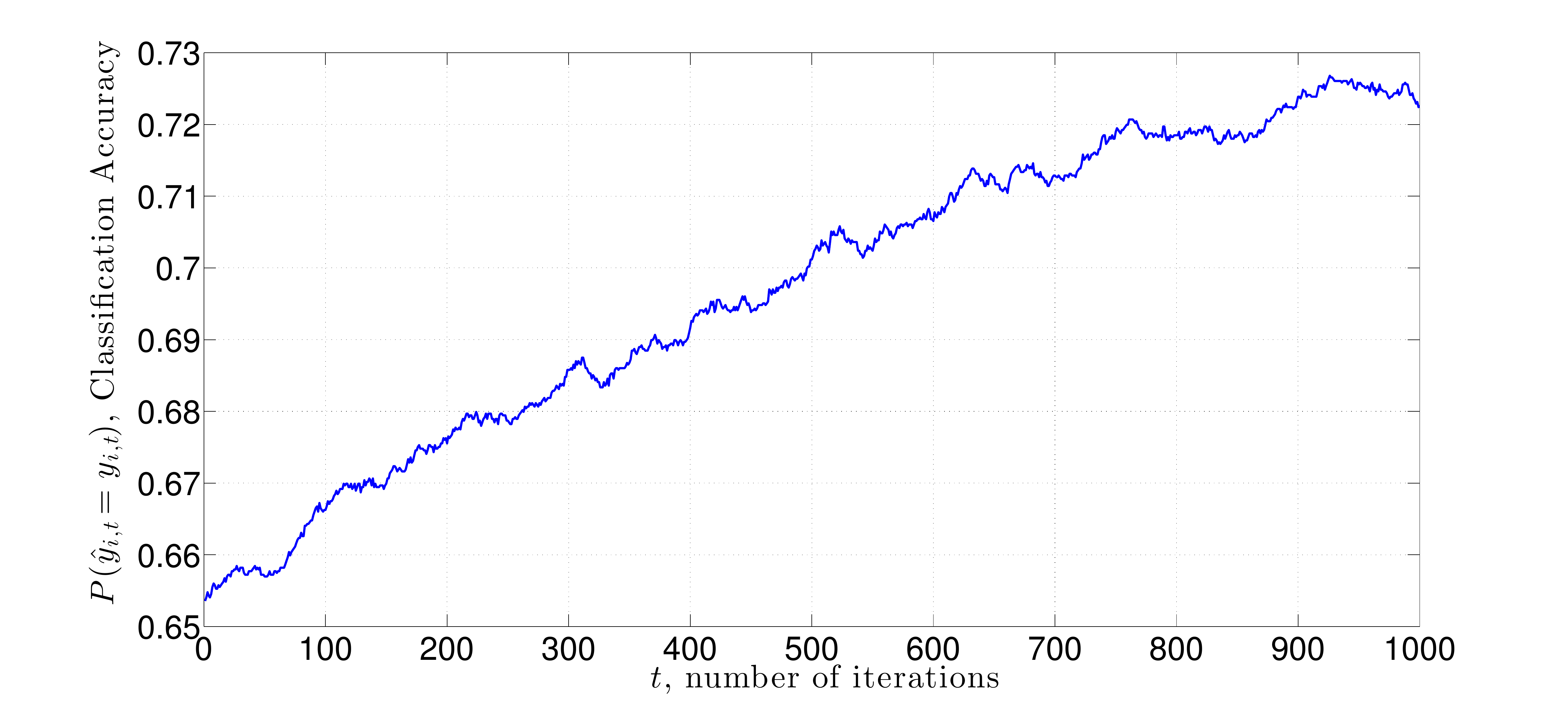}
	\caption{D4L classifier accuracy on the IRA texture dataset with step-size $\epsilon=5\times10^{-3}$. In a decentralized online manner, a robotic network is allowed to learn to make predictive inferences regarding the navigability differences between grass and concrete.}
	\label{fig:dvc_ira3small}	\vspace{-6mm}
\end{figure}

Note that each robot in the network only has access to information about the environment based on the path it has traversed, which may omit regions of the feature space crucial for recognizing patterns. By communicating with other robots in the network, individuals may learn over a broader domain associated with that which has been explored by the whole mobile network, and thus more effectively predict unexpected maneuvers due to the operating environment.
%
We select $\eps=5\times10^{-3}$ due to the challenging nature of the empirical distribution of robotic observations. However, such a small step-size makes learning occur at a slow rate. 
We repeat our field experiments $\bar{T}=10$ times and report the empirical mean of the results. We display this result in Figure \ref{fig:dvc_ira3small} for the field implementation of a $N=3$ robotic network. Observe that learning occurs, albeit slowly: the accuracy of the algorithm continues to climb as more data is accumulated, though by $T=10^3$ classification accuracy approaches only $73\%$.

%
%
%
%

\vspace{-2mm}
\section{Conclusion} \label{sec:conclusion}

This work represents the first attempt to extend the discriminative dictionary learning problem of \cite{10.1109/TPAMI.2011.156}  to networked settings. To do so we consider cases where losses are node-separable and introduced agreement constraints, yielding a decentralized stochastic non-convex optimization problem. By considering the Lagranian relaxation of an agreement-constrained system, we develop a block variant of the Arrow-Hurwicz saddle point method to solve it. Moreover, we establish the  convergence of the algorithm to a KKT point of the problem in expectation.

Experiments on a texture classification problem demonstrated comparable classifier performance between the centralized and decentralized settings, and illustrated the convergence rate dependence on the network.  Moreover, the proposed method allows multi-agent systems to learn over a new class of pattern recognition problems. In doing so, networks of interconnected computing servers may collaboratively solve such problems at an accelerated rate as compared with centralized methods. 
We additionally applied this method to a mobile robotic network deployed in an unknown domain charged with the task of collaboratively analyzing the navigability of distinct paths traversed by each robot, such that the unexpected maneuvers made by one robot may be avoid by another. 
\vspace{-2mm}


\section*{Appendix A: Proof of Proposition \ref{prop}} \label{prop_proof}

The set $\ccalD^N$ may be written as a Cartesian product of sets $\ccalD$. We assume that projection $\ccalP_{\ccalD^N}[\cdot]$ of the stacked iterates $\bbD$ into $\ccalD^N$ is equivalent to the separate projection $\ccalP_{\ccalD}[\cdot]$ of the components node-wise components $\bbD_i$ into the sets $\ccalD$.
The other primal domain, as well as the dual domains, are defined as Cartesian products of lower dimensional sets for each node and edge.
We assume a similar condition holds for the set projections with onto the stacked primal and dual sets $\ccalW^N$, $\mathfrak{L}^M$, and $\ccalN^M$, allowing the stacked iterates to be separated into their local node and edge-wise components via projections onto local sets $\ccalW$, $\mathfrak{L}$, and $\ccalN$.

To compute the primal stochastic gradient of the Lagrangian [cf. \eqref{eq:lagrangian}] with respect to a local dictionary $\bbD_{i,t}$ for a signal-output pair $(\bbx_{i,t}, \bby_{i,t})$, apply the node-separability of the global cost in \eqref{eq:costsum} to the first term in \eqref{eq:lagrangian}, and note that all terms of the derivative of the second term with respect to $\bbD_i$ in \eqref{eq:lagrangian} are null except those associated with node $i$ and neighbors $j$, to obtain
\begin{align}  \label{eq:dict_grad_appendix}\vspace{-1mm}
   \nabla_{\bbD_i} \hat{\ccalL}_t (\bbD_t,\bbw_t, \bbLam_t,\bbnu_t) &=
 	 \nabla_{\bbD_i} h_i (\bbD_{i,t}, \bbw_{i,t} ; (\bbx_{i,t}, \bby_{i,t}) ) \nonumber \\
	 & \quad \!+ \sum_{j \in n_i} (\bbLam_{ij,t} -\bbLam_{ji,t}) 
\vspace{-1mm} \end{align}\vspace{-1mm}
Here $\tbD_{i,t}$ is an element of the normal cone of $\ccalD$ when $\bbD_{i,t}$ is near the boundary of the set, and null otherwise, and comes from the use of projected stochastic gradients rather than stochastic gradients. Substitute this into the update \eqref{eq:sp_dict} and separate update along direction associated with agent $i$ to obtain 
 \begin{align}\label{eq:dict_update_appendix}\vspace{-1mm}
    \bbD_{i,t+1} 
      & = \ccalP_{\ccalD} \Big[ \bbD_{i,t}  - \eps_t  
          \Big(  	 \nabla_{\bbD_i} h_i (\bbD_{i,t}, \bbw_{i,t} ; (\bbx_{i,t}, \bby_{i,t}) ) \nonumber \\
	 & \quad+ \sum_{j \in n_i} (\bbLam_{ij,t} -\bbLam_{ji,t}) \Big)\Big] \;, 
\end{align}
To compute the primal stochastic gradient of  \eqref{eq:lagrangian} with respect $\bbw_i$, we apply the same logic to terms involving $\bbw_i$ to write
\begin{align}\label{eq:model_grad_appendix}\vspace{-1mm}
       \nabla_{\bbw_i} \hat{\ccalL}_t (\bbD_t,\bbw_t, \bbLam_t,\bbnu_t) &=      \nabla_{\bbw_i} h_i (\bbD_{i,t}, \bbw_{i,t} ; (\bbx_{i,t}, \bby_{i,t}) )  \nonumber \\
&\quad  +  \sum_{j \in n_i} (\bbnu_{ij,t} -\bbnu_{ji,t})
\end{align}
Again, the stochastic gradient of agent $i$ is decoupled from all others except the Lagrange multipliers it receives from its neighbors, and hence yields a decentralized implementation. The particular form of the model parameter update follows
\begin{align} \label{eq:param_update_appendix}\vspace{-1mm}
   \bbw_{i,t+1} 
      & = \ccalP_{\ccalW} \Big[ \bbw_{i,t} - \eps_t  
         \Big(\nabla_{\bbw_i} h_i (\bbD_{i,t}, \bbw_{i,t} ; (\bbx_{i,t}, \bby_{i,t}) ) \nonumber\\
	 &\qquad+ \sum_{j \in n_i} (\bbnu_{ij,t} -\bbnu_{ji,t})\Big)\Big] \;.
\end{align}

To develop the dual variable updates, compute the stochastic subgradient of \eqref{eq:lagrangian} with respect to the Lagrange multipliers associated with edge $(i,j)$ and the dictionary agreement constraint. By noting that all terms in the sum $\text{tr}(\bbLambda^T \bbC \bbD)$ are null except those associated with edge $(i,j)$, we obtain
\begin{equation} \label{eq:dual_grad_lam_appendix}
   \nabla_{\bbLam_{ij}} \hat{\ccalL}_t  (\bbD_{t+1},\bbw_{t+1}, \bbLam_t,\bbnu_t)
      =   \bbD_{i,t+1} - \bbD_{j,t+1}  \; .
\end{equation}
This local subgradient  corresponds to the communication link between agent $i$ and agent $j$. Separating the update in \eqref{eq:sp_dual_w} along variables associated with edge $(i,j)$, we obtain the local update
\begin{align} \label{eq:sda_dual_lam_ij_appendix}
   \bbLam_{ij,t+1} =  \ccalP_{\mathfrak{L}} \Big[ \bbLam_{ij,t} + \eps_t \left( \bbD_{i,t+1} - \bbD_{j,t+1}\right)  \Big] \;.
\end{align}

With analogous reasoning regarding the agreement constraint slack term for $\bbw$ in \eqref{eq:lagrangian}, we obtain 
\begin{equation} \label{eq:dual_grad_nu_appendix}
   \nabla_{\bbnu_{ij}} \hat{\ccalL}_t (\bbD_{t+1},\bbw_{t+1}, \bbLam_t,\bbnu_t)
      =    \bbw_{i,t+1} - \bbw_{j,t+1}  \; .
\end{equation}
With the same reasoning, the dual update for the Lagrange multipliers associated with the model parameter agreement constraints may be expressed as
\begin{align} \label{eq:sda_dual_nu_ij_appendix}
   \bbnu_{ij,t+1} =\ccalP_{\ccalN} \Big[ \bbnu_{ij,t} + \eps_t \left( \bbw_{i,t} - \bbw_{j,t} \right) \Big]  \;.
\end{align}
which is as stated in Proposition \ref{prop}.\vspace{-2mm}


\begin{thebibliography}{10}
\providecommand{\url}[1]{#1}
\csname url@samestyle\endcsname
\providecommand{\newblock}{\relax}
\providecommand{\bibinfo}[2]{#2}
\providecommand{\BIBentrySTDinterwordspacing}{\spaceskip=0pt\relax}
\providecommand{\BIBentryALTinterwordstretchfactor}{4}
\providecommand{\BIBentryALTinterwordspacing}{\spaceskip=\fontdimen2\font plus
\BIBentryALTinterwordstretchfactor\fontdimen3\font minus
  \fontdimen4\font\relax}
\providecommand{\BIBforeignlanguage}[2]{{%
\expandafter\ifx\csname l@#1\endcsname\relax
\typeout{** WARNING: IEEEtran.bst: No hyphenation pattern has been}%
\typeout{** loaded for the language `#1'. Using the pattern for}%
\typeout{** the default language instead.}%
\else
\language=\csname l@#1\endcsname
\fi
#2}}
\providecommand{\BIBdecl}{\relax}
\BIBdecl

\bibitem{cKoppelEtal15a}
A.~Koppel, G.~Warnell, E.~Stump, and A.~Ribeiro, ``D4l: Decentralized dynamic
  discriminative dictionary learning,'' in \emph{Proc. Int. Conf. Intelligent
  Robots, Systems}.\hskip 1em plus 0.5em minus 0.4em\relax Hamburg, Germany,
  September 28 - October 2 2015.

\bibitem{cKoppelEtal15b}
A.~Koppel, G.~Warnell, and E.~Stump, ``A stochastic primal-dual algorithm for
  task-driven dictionary learning in networks,'' in \emph{Asilomar Conf. on
  Signals, Systems, and Computers}, vol. (to appear).\hskip 1em plus 0.5em
  minus 0.4em\relax Pacific Grove, CA, November 8 - 11 2015.

\bibitem{Chen98atomicdecomposition}
S.~S. Chen, D.~L. Donoho, Michael, and A.~Saunders, ``Atomic decomposition by
  basis pursuit,'' \emph{SIAM Journal on Scientific Computing}, vol.~20, pp.
  33--61, 1998.

\bibitem{Jolliffe:1986}
I.~Jolliffe, \emph{Principal Component Analysis}.\hskip 1em plus 0.5em minus
  0.4em\relax Springer Verlag, 1986.

\bibitem{Mallat:2008:WTS:1525499}
S.~Mallat, \emph{A Wavelet Tour of Signal Processing, Third Edition: The Sparse
  Way}, 3rd~ed.\hskip 1em plus 0.5em minus 0.4em\relax Academic Press, 2008.

\bibitem{Elad:2006:IDV:2319066.2321231}
\BIBentryALTinterwordspacing
M.~Elad and M.~Aharon, ``Image denoising via sparse and redundant
  representations over learned dictionaries,'' \emph{Trans. Img. Proc.},
  vol.~15, no.~12, pp. 3736--3745, Dec. 2006. [Online]. Available:
  \url{http://dx.doi.org/10.1109/TIP.2006.881969}
\BIBentrySTDinterwordspacing

\bibitem{Mairal07sparserepresentation}
J.~Mairal, J.~Mairal, M.~Elad, M.~Elad, G.~Sapiro, and G.~Sapiro, ``Sparse
  representation for color image restoration,'' in \emph{the IEEE Trans. on
  Image Processing}.\hskip 1em plus 0.5em minus 0.4em\relax ITIP, 2007, pp.
  53--69.

\bibitem{Zibulevsky:2001:BSS:1120451.1120461}
\BIBentryALTinterwordspacing
M.~Zibulevsky and B.~A. Pearlmutter, ``Blind source separation by sparse
  decomposition in a signal dictionary,'' \emph{Neural Comput.}, vol.~13,
  no.~4, pp. 863--882, Apr. 2001. [Online]. Available:
  \url{http://dx.doi.org/10.1162/089976601300014385}
\BIBentrySTDinterwordspacing

\bibitem{Raina:2007:SLT:1273496.1273592}
\BIBentryALTinterwordspacing
R.~Raina, A.~Battle, H.~Lee, B.~Packer, and A.~Y. Ng, ``Self-taught learning:
  Transfer learning from unlabeled data,'' in \emph{Proceedings of the 24th
  International Conference on Machine Learning}, ser. ICML '07.\hskip 1em plus
  0.5em minus 0.4em\relax New York, NY, USA: ACM, 2007, pp. 759--766. [Online].
  Available: \url{http://doi.acm.org/10.1145/1273496.1273592}
\BIBentrySTDinterwordspacing

\bibitem{Wright:2009:RFR:1495801.1496037}
\BIBentryALTinterwordspacing
J.~Wright, A.~Y. Yang, A.~Ganesh, S.~S. Sastry, and Y.~Ma, ``Robust face
  recognition via sparse representation,'' \emph{IEEE Trans. Pattern Anal.
  Mach. Intell.}, vol.~31, no.~2, pp. 210--227, Feb. 2009. [Online]. Available:
  \url{http://dx.doi.org/10.1109/TPAMI.2008.79}
\BIBentrySTDinterwordspacing

\bibitem{10.1109/TPAMI.2011.156}
F.~Bach, J.~Mairal, and J.~Ponce, ``Task-driven dictionary learning,''
  \emph{IEEE Transactions on Pattern Analysis and Machine Intelligence},
  vol.~34, no.~4, pp. 791--804, 2012.

\bibitem{conf/cvpr/MairalBPSZ08}
\BIBentryALTinterwordspacing
J.~Mairal, F.~Bach, J.~Ponce, G.~Sapiro, and A.~Zisserman, ``Discriminative
  learned dictionaries for local image analysis.'' in \emph{CVPR}.\hskip 1em
  plus 0.5em minus 0.4em\relax IEEE Computer Society, 2008. [Online].
  Available:
  \url{http://dblp.uni-trier.de/db/conf/cvpr/cvpr2008.html#MairalBPSZ08}
\BIBentrySTDinterwordspacing

\bibitem{Bradley_2008_6192}
D.~Bradley and Bagnell, ``{Differentiable Sparse Coding},'' in
  \emph{Proceedings of Neural Information Processing Systems 22}, Dec. 2008.

\bibitem{conf/cvpr/JiangLD11}
\BIBentryALTinterwordspacing
Z.~Jiang, Z.~Lin, and L.~S. Davis, ``Learning a discriminative dictionary for
  sparse coding via label consistent k-svd.'' in \emph{CVPR}.\hskip 1em plus
  0.5em minus 0.4em\relax IEEE, 2011, pp. 1697--1704. [Online]. Available:
  \url{http://dblp.uni-trier.de/db/conf/cvpr/cvpr2011.html#JiangLD11}
\BIBentrySTDinterwordspacing

\bibitem{Tseng01convergenceof}
P.~Tseng and C.~O.~L. Mangasarian, ``Convergence of a block coordinate descent
  method for nondifferentiable minimization,'' \emph{J. Optim Theory Appl}, pp.
  475--494, 2001.

\bibitem{DBLP:journals/npsc/BezdekH03}
\BIBentryALTinterwordspacing
J.~C. Bezdek and R.~J. Hathaway, ``Convergence of alternating optimization,''
  \emph{Neural Parallel {\&} Scientific Comp.}, vol.~11, no.~4, pp. 351--368,
  2003. [Online]. Available: \url{http://dl.acm.org/citation.cfm?id=964886}
\BIBentrySTDinterwordspacing

\bibitem{xu2014globally}
Y.~Xu and W.~Yin, ``A globally convergent algorithm for nonconvex optimization
  based on block coordinate update,'' \emph{arXiv preprint arXiv:1410.1386},
  2014.

\bibitem{Mairal:2010:OLM:1756006.1756008}
\BIBentryALTinterwordspacing
J.~Mairal, F.~Bach, J.~Ponce, and G.~Sapiro, ``Online learning for matrix
  factorization and sparse coding,'' \emph{J. Mach. Learn. Res.}, vol.~11, pp.
  19--60, Mar. 2010. [Online]. Available:
  \url{http://dl.acm.org/citation.cfm?id=1756006.1756008}
\BIBentrySTDinterwordspacing

\bibitem{robbins1951}
\BIBentryALTinterwordspacing
H.~Robbins and S.~Monro, ``A stochastic approximation method,'' \emph{Ann.
  Math. Statist.}, vol.~22, no.~3, pp. 400--407, 09 1951. [Online]. Available:
  \url{http://dx.doi.org/10.1214/aoms/1177729586}
\BIBentrySTDinterwordspacing

\bibitem{Nedic09subgradientmethods}
A.~Nedic and A.~Ozdaglar, ``Subgradient methods for saddle-point problems,''
  \emph{J Optimiz. Theory App.}, vol. 142, no.~1, pp. 205--228, Aug. 2009.

\bibitem{DBLP:journals/corr/abs-1112-2972}
D.~Jakovetic, J.~M.~F. Xavier, and J.~M.~F. Moura, ``Fast distributed gradient
  methods,'' \emph{CoRR}, vol. abs/1112.2972, Apr. 2011.

\bibitem{RamNedicVeeravalli}
S.~Ram, A.~Nedic, and V.~Veeravalli, ``Distributed stochastic subgradient
  projection algorithms for convex optimization,'' \emph{J Optimiz. Theory
  App.}, vol. 147, no.~3, pp. 516--545, Sep. 2010.

\bibitem{yuan2013convergence}
K.~Yuan, Q.~Ling, and W.~Yin, ``On the convergence of decentralized gradient
  descent,'' \emph{ArXiv e-prints 1310.7063}, Oct. 2013.

\bibitem{chainais2013learning}
P.~Chainais and C.~Richard, ``Learning a common dictionary over a sensor
  network,'' in \emph{Computational Advances in Multi-Sensor Adaptive
  Processing (CAMSAP), 2013 IEEE 5th International Workshop on}.\hskip 1em plus
  0.5em minus 0.4em\relax IEEE, 2013, pp. 133--136.

\bibitem{1506308}
M.~Rabbat, R.~Nowak, and J.~Bucklew, ``Generalized consensus computation in
  networked systems with erasure links,'' in \emph{IEEE 6th Workshop Signal
  Process. Adv. in Wireless Commun Process.}, Jun. 5-8 2005, pp. 1088--1092.

\bibitem{JakubiecEtal13}
F.~Jakubiec and A.~Ribeiro, ``D-map: Distributed maximum a posteriori
  probability estimation of dynamic systems,'' \emph{IEEE Trans. Signal
  Process.}, vol.~61, no.~2, pp. 450--466, Feb. 2013.

\bibitem{ArrHurUza58}
K.~Arrow, L.~Hurwicz, and H.~Uzawa, \emph{Studies in linear and non-linear
  programming}, ser. Stanford Mathematical Studies in the Social
  Sciences.\hskip 1em plus 0.5em minus 0.4em\relax Stanford University Press,
  Stanford, Dec. 1958, vol.~II.

\bibitem{DBLP:dblp_conf/icassp/KoppelJR14}
A.~Koppel, F.~Y. Jakubiec, and A.~Ribeiro, ``A saddle point algorithm for
  networked online convex optimization.'' in \emph{Proc. Int. Conf. Acoust.
  Speech Signal Process.}, May 4-9 2014, pp. 8292--8296.

\bibitem{Koppelelat14}
A.~Koppel, F.~Jakubeic, and A.~Ribeiro, ``A saddle point algorithm for
  networked online convex optimization,'' \emph{IEEE Trans. Signal Process.},
  p.~14, Oct. 2014, available at {\footnotesize
  {http://www.seas.upenn.edu/~aribeiro/wiki}}.

\bibitem{aharon2006svd}
M.~Aharon, M.~Elad, and A.~Bruckstein, ``-svd: An algorithm for designing
  overcomplete dictionaries for sparse representation,'' \emph{Signal
  Processing, IEEE Transactions on}, vol.~54, no.~11, pp. 4311--4322, 2006.

\bibitem{Jenatton2009}
R.~Jenatton, J.-Y. Audibert, and F.~Bach, ``Structured variable selection with
  sparsity-inducing norms,'' Preprint arXiv:0904.3523, Tech. Rep., 2009.

\bibitem{bach2011convex}
F.~Bach, R.~Jenatton, J.~Mairal, and G.~Obozinski, ``Convex optimization with
  sparsity-inducing norms,'' in \emph{Optimization for Machine Learning}.\hskip
  1em plus 0.5em minus 0.4em\relax MIT Press, 2011.

\bibitem{Ber99}
D.~P. Bertsekas, \emph{Nonlinear Programming}.\hskip 1em plus 0.5em minus
  0.4em\relax Belmont, MA: Athena Scientific, 1999.

\bibitem{Birgin00nonmonotonespectral}
E.~G. Birgin, J.~É.~M. Martínez, and M.~Raydan, ``Nonmonotone spectral
  projected gradient methods on convex sets,'' \emph{SIAM Journal on
  Optimization}, pp. 1196--1211, 2000.

\bibitem{Efron04leastangle}
B.~Efron, T.~Hastie, I.~Johnstone, and R.~Tibshirani, ``Least angle
  regression,'' \emph{Annals of Statistics}, vol.~32, pp. 407--499, 2004.

\bibitem{Brodatz1966}
P.~Brodatz, \emph{Textures: A Photographic Album for Artists and
  Designers}.\hskip 1em plus 0.5em minus 0.4em\relax Dover, 1966.

\bibitem{Leung1999}
T.~Leung and J.~Malik, ``{Representing and Recognizing the Visual Appearence of
  Materials using Three-dimensional Textons},'' \emph{International Journal of
  Computer Vision}, vol.~43, no.~1, pp. 29--44, 1999.

\bibitem{Murphy2012}
K.~Murphy, \emph{Machine Learning: A Probabilistic Perspective}.\hskip 1em plus
  0.5em minus 0.4em\relax MIT press, 2012.

\bibitem{watts}
D.~J. Watts and S.~H. Strogatz, ``Collective dynamics of 'small-world'
  networks,'' \emph{Nature}, vol. 393, no. 6684, pp. 440--442, Jun. 1998.

\bibitem{west_introduction_2000}
D.~B. West, \emph{Introduction to Graph Theory}, 2nd~ed.\hskip 1em plus 0.5em
  minus 0.4em\relax Prentice Hall, Sept. 2000.

\bibitem{cKoppelEtal16a}
A.~Koppel, G.~Warnell, J.~Fink, E.~Stump, and A.~Ribeiro, ``Online learning for
  characterizing unknown environments in ground robotic vehicle models,'' in
  \emph{Proc. Int. Conf. Intelligent Robots and Systems}, vol. (submitted).\hskip 1em
  plus 0.5em minus 0.4em\relax Daejeon, Korea, Oct 9 - 14 2016.

\end{thebibliography}
   \end{document}